\documentclass[pmlr]{jmlr}

\usepackage{longtable}
\usepackage{multirow}
\usepackage{algorithm}
\usepackage{algpseudocode}
\usepackage{amsmath}
\usepackage{amssymb}
 \usepackage{booktabs}
 
\usepackage[load-configurations=version-1]{siunitx} 

\makeatletter
\def\set@curr@file#1{\def\@curr@file{#1}} 
\makeatother

\theorembodyfont{\upshape}
\theoremheaderfont{\scshape}
\theorempostheader{:}
\theoremsep{\newline}

\jmlrvolume{219}
\jmlryear{2023}
\jmlrworkshop{Machine Learning for Healthcare}

\title{Efficient Representation Learning for Healthcare with Cross-Architectural Self-Supervision}

\author{\Name{Pranav Singh}
       \Email{ps4364@nyu.edu}\\ 
       \addr Department of Computer Science, Tandon School of Engineering\\
       New York University\\
       New York, NY 11202, USA \\
       \AND
       \Name{Jacopo Cirrone}
       \Email{cirrone@courant.nyu.edu}\\ 
       \addr Center for Data Science 
        and Colton Center for Autoimmunity\\
        New York University\\
        New York, NY 10011} 

\editor{Editor's name}

\begin{document}

\maketitle

\begin{abstract}
  In healthcare and biomedical applications, extreme computational requirements pose a significant barrier to adopting representation learning. Representation learning can enhance the performance of deep learning architectures by learning useful priors from limited medical data. However, state-of-the-art self-supervised techniques suffer from reduced performance when using smaller batch sizes or shorter pretraining epochs, which are more practical in clinical settings. We present Cross Architectural - Self Supervision (CASS) in response to this challenge. This novel siamese self-supervised learning approach synergistically leverages Transformer and Convolutional Neural Networks (CNN) for efficient learning.
Our empirical evaluation demonstrates that CASS-trained CNNs and Transformers outperform existing self-supervised learning methods across four diverse healthcare datasets. With only 1\% labeled data for finetuning, CASS achieves a 3.8\% average improvement; with 10\% labeled data, it gains 5.9\%; and with 100\% labeled data, it reaches a remarkable 10.13\% enhancement. Notably, CASS reduces pretraining time by 69\% compared to state-of-the-art methods, making it more amenable to clinical implementation. We also demonstrate that CASS is considerably more robust to variations in batch size and pretraining epochs, making it a suitable candidate for machine learning in healthcare applications.
\end{abstract}

\section{Introduction}
\label{introduction}
The application of artificial intelligence in medical imaging has been restricted by minimal data availability. First, data labeling typically requires domain-specific knowledge. Therefore, the requirement of large-scale clinical supervision may be cost and time prohibitive. Second, due to patient privacy, disease prevalence, and other limitations, it is often difficult to release imaging datasets for secondary analysis, research, and diagnosis. Third, due to an incomplete understanding of diseases. This could be either because the disease is emerging or because no mechanism is in place to systematically collect data about the prevalence and incidence of the disease. Due to the latter two conditions, even unlabeled data is limited. An example of the former is COVID-19, when despite collecting chest X-ray data spanning decades, the samples lacked data for COVID-19 (\cite{sriram2021covid}), while an example of the latter is autoimmune diseases. Statistically, autoimmune diseases affect 3\% of the US population or 9.9 million US citizens. There are still major outstanding research questions for autoimmune diseases regarding the presence of different cell types and their role in inflammation at the tissue level. Studying autoimmune diseases is critical because autoimmune diseases affect a large part of society, and these conditions have been on the rise recently (\cite{galeotti2020autoimmune,lerner2015world,ehrenfeld2020covid}). Other fields like cancer and MRI image analysis have benefited from the application of artificial intelligence (AI). To mitigate this problem, transfer learning has been the de facto solution. In transfer learning, an architecture trained on natural imaging datasets, such as ImageNet, is utilized and then fine-tuned on the medical imaging dataset. The emergence of other powerful paradigms, such as self-supervised learning, provides a better alternative. Pretraining with self-supervised techniques is label-free, wherein a deep neural network learns useful priors by solving a pre-training task on unlabeled images. This is especially useful in fields with limited labeled data availability or if the cost and effort required to provide annotations are high. With the help of these valuable and transferable priors, self-supervised training  can benefit from applying self-supervised techniques. 

But existing representation/self-supervised techniques are highly compute-intensive and require multiple GPU servers over multiple days. This limits their applicability, making them inaccessible to the general practitioner. One way to mitigate this issue is by reducing the number of pre-training epochs. During self-supervised pre-training, the deep learning model solves an auxiliary task such as image reconstruction and similarity/dissimilarity optimization in a label-free manner \cite{chen2020simple}. By solving the auxiliary task (without labels), the model learns helpful priors that it can reuse during the downstream task to perform better at that task.
But even the state-of-the-art self-supervised techniques suffer a marked reduction in performance when trained for fewer pre-training epochs. These techniques are pre-trained with a large batch size of 1024 or larger \cite{caron2021emerging,he2022masked,grill2020bootstrap}. In some cases, these batch sizes are several times larger than the size of the entire medical imaging dataset. Applying self-supervised techniques with large batch sizes is particularly challenging due to minimal data availability (labeled and unlabeled images). For example, the median dataset size for autoimmune diseases is 99-540 samples  \cite{tsakalidou2022computer, Stafford2020ASR}. Due to these extremely small sizes, sometimes it is impossible to run large batch sizes. Existing self-supervised techniques drop a significant performance when used with small batch sizes. For example, state-of-the-art DINO \cite{caron2021emerging} drops classification performance by as much as 25\% when trained with a reduced batch size of 8.

To address these issues and improve the applicability of self-supervised learning for medical imaging, we propose to combine CNN and Transformer in a response-based siamese contrastive method. With this approach, the extracted representations of each input image are compared across two branches representing each architecture (see Figure \ref{fig:CASS}). By transferring features sensitive to translation equivariance and locality from CNN to Transformer, our proposed approach - CASS, learns more predictive data representations in limited data scenarios where a Transformer-only model cannot find them. We studied this quantitatively and qualitatively in Section \ref{results}. 
\subsection*{Generalizable Insights about Machine Learning in the Context of Healthcare}
\begin{itemize}
\item We introduce \textbf{C}ross \textbf{A}rchitectural  - \textbf{S}elf \textbf{S}upervision (CASS), a siamese CNN-Transformer approach for learning improved data representations in a self-supervised setting through architecture invariance instead of augmentation invariance. We study the efficacy of our approach in limited data availability in the medical image analysis domain \footnote{We have open-sourced our code at \url{https://github.com/pranavsinghps1/CASS}}
\item We propose the use of CASS for analysis of autoimmune diseases such as dermatomyositis and demonstrate an improvement of 2.55\% 
compared to the existing state-of-the-art self-supervised approaches. To our knowledge, the autoimmune dataset contains 198 images and is the smallest known dataset studied in self-supervised learning.   
\item As most of the existing self-supervised techniques have been studied on natural imaging datasets, their study and applicability in medical imaging has been limited. We evaluate CASS and DINO on three challenging medical image analysis problems (autoimmune disease cell classification, brain tumor classification, and skin lesion classification) on three public datasets (Dermofit Project Dataset \cite{Dermofit}, brain tumor MRI Dataset \cite{Cheng2017,s21062222} and ISIC 2019~\cite{Tschandl2018TheHD, Gutman2018SkinLA, Combalia2019BCN20000DL}) and find that CASS improves classification performance (F1 Score and Recall value) over the existing state of the art self-supervised techniques by an average of ~3.8\% using 1\% label fractions, 5.9 \% with 10\% label fractions and 10.13\% with 100\% label fractions. We also compare CASS with an additional contrastive technique BYOL \cite{Grill2020BootstrapYO}  and a reconstructive self-supervised technique MAE \cite{he2022masked}.
\item Existing methods also suffer a severe drop in performance when trained for a reduced number of epochs or batch size (\cite{caron2021emerging, Grill2020BootstrapYO, chen2020simple}). We show that CASS is robust to these changes in Sections \ref{epoch-var}.
\item New state-of-the-art self-supervised techniques often require significant computational requirements. This is a major hurdle as these methods can take around 20 GPU days to train \cite{Azizi2021BigSM}. This makes them inaccessible in limited computational resource settings. CASS, on average, takes 69\% less time than the existing state-of-the-art methods. We further expand on this result in Section \ref{results}.
\end{itemize}

\section{Related Work}

\subsection{Neural Network Architectures for Image Analysis}
\label{archs}
CNNs are a famous architecture of choice for many image analysis applications~\cite{khan2020survey}. CNNs learn more abstract visual concepts with a gradually increasing receptive field. They have two favorable inductive biases: (i) translation equivariance resulting in the ability to learn equally well with shifted object positions, and (ii) locality resulting in the ability to capture pixel-level closeness in the input data. CNNs have been used for many medical image analysis applications, such as disease diagnosis~\cite{yadav2019deep} or semantic segmentation~\cite{ronneberger2015u}. To address the requirement of additional context for a more holistic image understanding, the Vision Transformer (ViT) architecture \cite{dosovitskiy2020image} has been adapted to images from language-related tasks and recently gained popularity~\cite{liu2021Swin,liu2021swinv2,pmlr-v139-touvron21a}. In a ViT, the input image is split into patches treated as tokens in a self-attention mechanism. Compared to CNNs, ViTs can capture additional image context but lack ingrained inductive biases of translation and location. While CNNs have these inductive biases as a prior, Transformers learn them over training. As a result, ViTs typically outperform CNNs on larger datasets~\cite{d2021convit}. 

\subsubsection{Cross-architecture Technqiues}
As CNNs and Transformers both have their own sets of pros and cons. Multiple cross-architecture techniques have been proposed with the aim of combining the features of CNNs and Transformers; they can be classified into two categories (i) Hybrid cross-architecture techniques and (ii) Siamese cross-architecture techniques. 
Hybrid cross-architecture techniques combine parts of CNNs and Transformers in some capacity for a single model, allowing architectures to learn unique representations. ConViT~\cite{d2021convit} combines CNNs and ViTs using gated positional self-attention (GPSA) to create a soft convolution similar to inductive bias and improve upon the
capabilities of Transformers alone. More recently, the training regimes and inferences from ViTs have been used to design a new family of convolutional architectures - ConvNext \cite{liu2022convnet}, outperforming benchmarks set by ViTs in classification tasks. \cite{li2021bossnas} further simplified the procedure to create an optimal CNN-Transformer using their self-supervised Neural Architecture Search (NAS) approach. 

On the other hand, siamese cross-architecture techniques combine CNNs and Transformers without any changes to their architecture to help both of them learn better representations. \cite{Gong2022CMKDCC} used CNN and Transformer pairs in a consistent teaching knowledge distillation format for audio classification and showed that cross-architecture distillation makes distilled models less prone to overfitting and also improves robustness. Compared with the CNN-attention hybrid
models, cross-architecture knowledge distillation is more effective
and requires no model architecture change. Similarly, \cite{guo2022cross} also used a 3D-CNN and Transformer to learn strong representations and proposed a self-supervised learning module to predict an edit distance between two video sequences in the temporal order.

\subsection{Self-Supervised Learning}
\label{SSL-intro}
Most existing self-supervised techniques can be classified into contrastive and reconstruction-based techniques. Traditionally, contrastive self-supervised techniques have been trained by optimizing the distance between representations of different augmented views of the same image (‘positive pairs’) and representations of augmented views from different images (‘negative pairs’) \cite {He2020MomentumCF,Chen2020ASF,caron2020unsupervised}. But this is highly memory intensive as we need to track positive and negative pairs. Recently, Bootstrap Your Own Latent (BYOL) \cite{Grill2020BootstrapYO} and DINO \cite{caron2021emerging} have improved upon this approach by eliminating the memory banks. The premise of using negative pairs is to avoid collapse. Several strategies have been developed with BYOL using a momentum encoder, Simple Siamese (SimSiam) \cite{Chen2021ExploringSS} a stop gradient, and DINO applying the counterbalancing effects of sharpening and centering to avoid collapse. Techniques relying only on positive pairs are much more efficient than the ones using positive and negative pairs. Recently, there has been a surge in reconstruction-based self-supervised pretraining methods with the introduction of MSN \cite{assran2022masked} and MAE \cite{he2022masked}. These methods aim to learn semantic knowledge of the image by masking some part of it and then predicting the masked portion during pretraining.

\subsubsection{Self-supervised Learning and Medical Image Analysis}
\label{ssl-med}
As mentioned in Section \ref{introduction}, natural image datasets like ImageNet are commonly used for benchmarking and comparing self-supervised techniques. But most natural datasets are balanced and have sizeable inter-class variability. This is not representative of real-world data, especially in medical imaging, characterized by class imbalance and slight inter-class variance \cite{cassidy2022analysis}. To improve their performance, some self-supervised methods use batch-level statistics that help them when applied to balanced datasets but limit their performance on unbalanced datasets. Batch statistic-dependent methods have been found to drop significant performance for image classification tasks when trained by artificially inducing class imbalance \cite{assran2022hidden}.
This prior of some self-supervised techniques like MSN \cite{assran2022masked}, SimCLR \cite{chen2020simple}, and VICreg \cite{bardes2021vicreg} limits their applicability on imbalanced datasets, especially in the case of medical imaging. 

Existing self-supervised techniques typically require large batch sizes and datasets to learn meaningful priors. When these conditions are not met, a marked reduction in performance is demonstrated~\cite{caron2021emerging,chen2020simple, Caron2020UnsupervisedLO, Grill2020BootstrapYO}. Self-supervised learning approaches are practical in big data medical applications~\cite{ghesu2022self,azizi2021big}, such as analysis of dermatology and radiology imaging. In more limited data scenarios (3,662 images - 25,333 images), \cite{Matsoukas2021IsIT} reported that ViTs outperform their CNN counterparts when self-supervised pre-training is followed by supervised fine-tuning. Transfer learning favors ViTs when applying standard
training protocols and settings. Their study included running the DINO \cite{caron2021emerging} self-supervised method over 300 epochs with a batch size of 256 with ImageNet initialization. However, questions remain about the accuracy and
efficiency of using existing self-supervised techniques on datasets whose entire size is smaller than their peak performance batch size.
In some instances, gathering sufficient unlabeled data for pertaining is impossible for reasons mentioned in Section \ref{introduction}. For example, in the case of  autoimmune diseases, due to a lack of existing pipelines and the rarity of the diseases \cite{singh2023data}, even unlabelled data is unavailable for pre-training; this restricts the applicability of the existing self-supervised methods. Also, since existing self-supervised methods are extremely computation intensive, from a general practitioner's perspective with limited computational power raises the
concern of how these methods could be leveraged for their practical application. While transfer learning provides a solution, pre-training with these takes can further improve performance on top of transfer learning. But for that, self-supervised techniques have to be more accessible. Adoption and faster development of self-supervised paradigms will only be possible when they become easy to plug and play with limited computational power.

In this work, we explore these questions by designing CASS, a novel self-supervised approach developed with the core values of efficiency and effectiveness. In simple terms, we combine CNN and Transformer in a response-based siamese contrastive method by reducing similarity to combine the abilities of CNNs and Transformers. This approach was initially designed for a 198-image dataset for muscle biopsies of inflammatory lesions from patients with dermatomyositis - an autoimmune disease. The benefits of this approach are illustrated by challenges in diagnosing autoimmune diseases due to their rarity, limited data availability, and heterogeneous features. Consequently, misdiagnoses are common, and the resulting diagnostic delay plays a significant factor in their high mortality rate. Autoimmune diseases share commonalities with COVID-19 regarding clinical manifestations, immune responses, and pathogenic mechanisms. Moreover, some patients have developed autoimmune diseases after COVID-19 infection \cite{Liu2020COVID19AA}. Despite this increasing prevalence, the representation of autoimmune diseases in medical imaging and deep learning is limited.

\section{Methods}

We start by motivating our method before explaining it in detail (in Section \ref{desc-cass}). Traditionally, contrastive self-supervised methods have used different augmentations of the same image to create positive pairs. These were then passed through the same architectures but with a different set of parameters \cite{Grill2020BootstrapYO}. In \cite{caron2021emerging}, the authors introduced image cropping of different sizes to add local and global information. They also used different operators and techniques to avoid collapse, as described in Section \ref{SSL-intro}.\\

Through architectural differences, we explore an alternative approach to creating positive pairs with CASS. 
\cite{Raghu2021DoVT} in their study suggested that for the same input, Transformers and CNNs extract different representations. They conducted their study by analyzing the CKA (Centered Kernel Alignment) for CNNs and Transformer using ResNet \cite{He2016DeepRL} and ViT (Vision Transformer) \cite{dosovitskiy2020image} family of encoders, respectively. They found that Transformers have a more uniform representation across all layers than CNNs. They also have self-attention, enabling global information aggregation from shallow layers and skip connections that connect lower layers to higher layers, promising information transfer. Hence, lower and higher layers in Transformers show much more similarity than in CNNs. The receptive field of lower layers for Transformers is more extensive than in CNNs. While this receptive field gradually grows for CNNs, it becomes global for Transformers around the midway point. Transformers don't attend locally in their earlier layers, while CNNs do. Using local information earlier is essential to restore intricate details. CNNs have a more centered receptive field than a more globally spread receptive field of Transformers. Hence, representations drawn from the same input will differ for Transformers and CNNs. Until now, self-supervised techniques have used only one kind of architecture at a time, either a CNN or Transformer. But differences in the representations learned with CNN and Transformers inspired us to create positive pairs using different architectures or feature extractors rather than a different set of augmentations. This, by design, avoids collapse as the two architectures will never give the exact representation as output. By contrasting their extracted features at the end, we hope to help the Transformer learn representations from CNN and vice versa. This should help both the architectures to learn better representations and learn from patterns that they would miss. We verify this by studying attention and feature maps from supervised and CASS-trained CNN and Transformers in Appendix \ref{attn-map-appendix} 

\subsection{Description of CASS}
\label{desc-cass}
CASS aims to extract and learn representations in a siamese self-supervised way similar to BYOL \cite{grill2020bootstrap} and DINO \cite{caron2021emerging}. But unlike BYOL and DINO, an image is passed through a common set of augmentations to achieve this. The augmented image is simultaneously passed through a CNN and Transformer to create positive pairs. The output logits from the CNN and Transformer are then used to find cosine similarity loss (equation \ref{loss_eq}). This is the same loss function as used in BYOL \cite{grill2020bootstrap}. Furthermore, the intuition of CASS is very similar to that of BYOL. In BYOL, the target and the online arm are differently parameterized to avoid collapse to a trivial solution, and an additional predictor is used with the online arm. They compared this setup to that of GANs, where the joint optimization of both arms to a typical value was impossible due to differences in the arms. Analogously, In CASS, instead of using an additional MLP on top of one of the arms and differently parameterizing them, we use two fundamentally different architectures. Since the two architectures give different output representations, as mentioned in \cite{Raghu2021DoVT}, the model doesn't collapse. 
We also report results for CASS using a different set of CNNs and Transformers in Appendix \ref{change-arch} and Section \ref{results}, and not a single case of the model collapse was registered. With this approach, we reduce the computational load by (i) reducing the number of times augmentations are applied. Instead of traditional contrastive techniques like BYOL and DINO, we apply augmentations only once. (ii) In DINO and BYOL, the student network's parameters are a lagging function of the teacher network's parameters; this requires these techniques to keep track of these parameters and then transfer them every epoch, while in CASS, both the arms are independent, and thus avoid this extra computation. 

\begin{equation}
\label{loss_eq}
\operatorname{loss} =2-2 \times
 \operatorname{F(R)} \times \operatorname{F(T)}
 \end{equation}
\begin{align*}
\text{where, }
\operatorname{F(x)}=\sum_{i=1}^{N} \left(\frac{x}{\left(\operatorname{max}\left(\|x\|_{2}\right), \epsilon\right)}\right) 
\end{align*}
In Equation \ref{loss_eq}, R and T represent embeddings from CNN and Transformer, respectively.
We use the same set of parameters for both architectures' optimizer and learning schedule. We also use stochastic weigh averaging (SWA) \cite{Izmailov2018AveragingWL} with Adam optimizer and a learning rate of 1e-3. For the learning rate, we use a cosine schedule with a maximum of 16 iterations and a minimum value of 1e-6. ResNets are typically trained with Stochastic Gradient Descent (SGD), and our use of the Adam optimizer is quite unconventional based on experiments by \cite{dosovitskiy2020image}. Furthermore, unlike existing self-supervised techniques, there is no parameter sharing between the two architectures. 



\begin{figure}[!htb]
    \centering
    \centering  
    \includegraphics[width=0.55\linewidth]{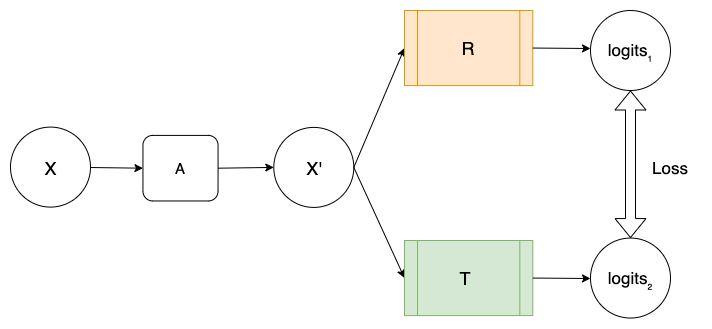}
    \hspace{1cm}
    \includegraphics[width=0.55\linewidth]{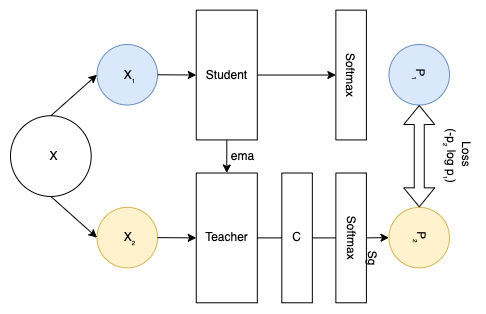}
    \caption{(Top) In our proposed self-supervised architecture - CASS, R represents ResNet-50, a CNN and T in the other box represents the Transformer used (ViT); X is the input image, which becomes X' after applying augmentations. Note that CASS applies only one set of augmentations to create X'. X' is passed through both arms to compute loss, as in Equation 1. This differs from DINO, which passes different augmentation of the same image through networks with the same architecture but different parameters. The output of the teacher network is centered on a mean computed over a batch. Another key difference is that in CASS, the loss is computed over logits; meanwhile, in DINO, it is computed over softmax output.}
\label{fig:CASS}
\end{figure}

In Figure~\ref{fig:CASS}, we show CASS on top and DINO \cite{caron2021emerging} at the bottom. Comparing the two, CASS does not use any extra mathematical treatment used in DINO to avoid collapse, such as centering and applying the softmax function on the output of its student and teacher networks. We also provide an ablation study using a softmax and sigmoid layer for CASS in Appendix \ref{softmax-appendix}. After training CASS and DINO for one cycle, DINO yields only one kind of trained architecture. In contrast, CASS provides two trained architectures (1 - CNN and 1 - Transformer). CASS-pre-trained architectures perform better than DINO-pre-trained architectures in most cases, as further elaborated in Section \ref{results}. 

\section{Experimental Details}

\subsection{Datasets}

Unless specified otherwise, we split the datasets into three splits - training, validation, and testing following the 70/10/20 split strategy. We expand on our rationale for datasets in Appendix \ref{ds-choice}, along with sample images from each of the datasets used.

\begin{table*}[!ht]
\centering
\begin{tabular}{llll}
\hline
\multicolumn{1}{c}{\multirow{2}{*}{Techniques}} & \multicolumn{1}{c}{\multirow{2}{*}{Backbone}} & \multicolumn{2}{l}{Testing F1 score} \\
\multicolumn{1}{c}{}                            & \multicolumn{1}{c}{}                           & 10\%       & 100\%         \\
\hline

DINO                                            & ResNet-50                                      & \textbf{0.8237±0.001}            &0.84252±0.008           \\
CASS                                           & ResNet-50                                   & 0.8158±0.0055           & \textbf{0.8650±0.0001}           \\
Transfer                                      & ResNet-50                                 & 0.819±0.0216          &  0.83895±0.007            \\
\hline

DINO                                            & ViT B/16                             & 0.8445±0.0008           &0.8639± 0.002             \\
CASS                                           & ViT B/16                                          & \textbf{0.8717±0.005}           & \textbf{0.8894±0.005}            \\
Transfer                                      & ViT B/16                                          &0.8356±0.007           &0.8420±0.009          \\
\hline
\end{tabular}
\caption{Results for autoimmune biopsy slides dataset. In this table, we compare the F1 score on the test set. We observed that CASS outperformed the existing state-of-art self-supervised method using 100\% labels for CNN and Transformers. Although DINO outperforms CASS for CNN with 10\% labeled fraction. Overall, CASS outperforms DINO by ~2.2\% for 100\% labeled training for CNN and Transformer. For Transformers in 10\% labeled training CASS' performance was ~2.7\% better than DINO.}
\label{ai-perf}
\end{table*}

\begin{table}[!htb]
\centering
\begin{tabular}{lll}
\hline
Dataset    & DINO              & CASS                    \\
\hline
Autoimmune & 1 H 13 M    & \textbf{21 M}          \\
Dermofit & 3 H 9 M & \textbf{1 H 11 M} \\
Brain MRI  & 26 H 21 M  & \textbf{7 H 11 M}  \\
ISIC-2019  & 109 H 21 M & \textbf{29 H 58 M} \\
\hline
\end{tabular}
\caption{Self-supervised pretraining time comparison for 100 epochs on a single RTX8000 GPU. In this table, H represents hour(s), and M represents minute(s).}
\label{computetime}
\end{table}


\paragraph{Autoimmune diseases biopsy slides (\cite{singh2023data, VANBUREN2022113233})} consists of slides cut from muscle biopsies of dermatomyositis patients stained with different proteins and imaged to generate a dataset of 198 TIFF images set from 7 patients. The presence or absence of these cells helps to diagnose dermatomyositis. Multiple cell classes can be present per image; this is a multi-label classification problem. 
    Our task here was to classify cells based on their protein staining into TFH-1, TFH-217, TFH-Like, B cells, and others. We used the F1 score as our metric for evaluation, as employed in previous works by \cite{singh2023data, VANBUREN2022113233}. These RGB images have a consistent size of 352 by 469.
    
\paragraph{Dermofit dataset (\cite{Dermofit})}  contains standard RGB images captured through an SLR camera indoors with ring lightning. There are 1300 image samples, classified into ten classes: Actinic Keratosis (AK), Basal Cell Carcinoma (BCC), Melanocytic Nevus / Mole (ML), Squamous Cell Carcinoma (SCC), Seborrhoeic Keratosis (SK), Intraepithelial carcinoma (IEC), Pyogenic Granuloma (PYO), Haemangioma (VASC), Dermatofibroma (DF) and  Melanoma (MEL). This dataset comprises images of different sizes; no two images are the same size. They range from 205×205 to 1020×1020 in size. Our pretext task is multi-class classification, and we use the F1 score as our evaluation metric on this dataset. 
    
\paragraph{Brain tumor MRI dataset (\cite{Cheng2017, Amin2022ANM, msoud_nickparvar_2021})} 7022 images of human brain MRI classified into four classes: glioma, meningioma, no tumor, and pituitary. 
The dataset curator have created predefined training and testing splits. We followed their splits, 5,712 images for training and 1,310 for testing. Since this was a combination of multiple datasets, the size of images varies throughout the dataset from 219×234 to 512×512. The pretext of the task is multi-class classification, and we used the F1 score as the metric. 
\paragraph{ISIC 2019 ({\cite{Tschandl2018TheHD, Gutman2018SkinLA, Combalia2019BCN20000DL}})} consists of 25,331 images across eight different categories - melanoma (MEL), melanocytic nevus (NV), Basal cell carcinoma (BCC), actinic keratosis(AK), benign keratosis(BKL), dermatofibroma(DF), vascular lesion (VASC) and Squamous cell carcinoma(SCC). This dataset contains images of size 600 × 450 and 1024 × 1024. The distribution of these labels is unbalanced across different classes with hard-to-distinguish classes. For evaluation, we followed the metric in the official competition, i.e.,  balanced multi-class accuracy value, which is semantically equal to recall. We provide a mathematical representation of all the metrics used in Appendix \ref{metrics}.

\subsection{Self-supervised learning}
We studied and compared results between DINO and CASS-pre-trained CNNs and Transformers. In addition to these, we also compared results with Bootstrap Your Latent BYOL \cite{grill2020bootstrap}, and Masked Auto-Encoders (MAE) \cite{he2022masked} on the autoimmune and Dermofit datasets in Section \ref{BYOL-cmp} and Appendix \ref{mae-cmp-appendix}, respectively.  For each experiment, we pre-trained for 100 epochs with a batch size of 16. We ran these experiments on an internal cluster with a single GPU unit (NVIDIA RTX8000) with 48 GB video RAM, 2 CPU cores, and 64 GB system RAM. We provide an extended list of Protocols and hyperparameters in Appendix \ref{ssl-train}.
\subsection{End-to-end fine-tuning} To evaluate the utility of the learned representations, we use the self-supervised pre-trained weights for the downstream classification tasks. While performing the downstream fine-tuning, we perform the entire model (E2E
fine-tuning). The test set metrics were used as proxies for representation quality. We trained the entire model for a maximum of 50 epochs with an early stopping patience of 5 epochs. For supervised fine-tuning, we used Adam optimizer with a cosine annealing learning rate starting at 3e-04. Since almost all medical datasets have some class imbalance, we applied class distribution normalized Focal Loss \cite{Lin2017FocalLF} to navigate class imbalance.

Additionally, we fine-tuned the models using different label fractions during E2E fine-tuning, i.e.,  1\%, 10\%, and 100\& label fractions. For example, if a model is trained with a 10\% label fraction, then that model will have access only to 10\% of the training dataset samples and their corresponding labels during the E2E
fine-tuning after initializing weights after pretraining. 

\section{Results and Discussion}
\label{results}

\subsection{Compute and Time analysis Analysis}
\label{time}
We ran all the experiments on a single NVIDIA
RTX8000 GPU with 48GB video memory. In Table \ref{computetime}, we compare the cumulative training times for a CNN and Transformer self-supervised training with DINO and CASS. We observed that CASS took an average of 69\% less time than DINO. 

\subsection{Results on the four medical imaging datasets}
We did not perform 1\% finetuning for the autoimmune diseases biopsy slides of 198 images because using 1\% images would be too small a number to learn anything meaningful, and the results would be highly randomized. Similarly, we did not perform 1\% fine-tuning for the dermofit dataset as the training set was too small to draw meaningful results with just ten samples. We averaged the results over five different runs with different seed values. We presented the results in a 95\% confidence interval on the four medical imaging datasets in Tables \ref{ai-perf}, \ref{dermofitperformance}, \ref{brainMRIperformance}, and \ref{ISICperformance}. We also compared the standard transfer learning approach (represented by Transfer in Tables \ref{ai-perf}, \ref{dermofitperformance}, \ref{brainMRIperformance}, and \ref{ISICperformance}) with the self-supervised approach. We observed that self-supervised pre-training performs better than transfer learning in all cases by a margin. We also observed that CASS improves upon the classification performance of existing state-of-the-art self-supervised method DINO by 3.8\% with
1\% labeled data, 5.9\% with 10\% labeled data, and 10.13\% with 100\% labeled data. We also compare BYOL and MAE in Section \ref{BYOL-cmp} and Section \ref{mae-cmp-appendix}, respectively. 

\begin{table*}[!htb]
\centering
\begin{tabular}{lll}

\hline
\multicolumn{1}{c}{\multirow{2}{*}{Techniques}} & 
\multicolumn{2}{l}{Testing F1 score} \\
\multicolumn{1}{c}{}                                         & 10\%       & 100\%         \\
\hline

DINO                                             (Resnet-50)                                     &0.3749±0.0011          &0.6775±0.0005           \\
CASS                                            (Resnet-50)                                     & \textbf{0.4367±0.0002}      & \textbf{0.7132±0.0003}           \\
Transfer                                       (Resnet-50)                                   & 0.33±0.0001           &  0.6341±0.0077           \\
\hline

DINO                        (ViT B/16)                             &0.332± 0.0002             &0.4810±0.0012            \\
CASS    (ViT B/16)                                  & \textbf{0.3896±0.0013}           & \textbf{0.6667±0.0002}            \\
Transfer        (ViT B/16)                               &0.299±0.002          &0.456±0.0077          \\
\hline
\end{tabular}
\caption{This table contains the results for the dermofit dataset. CASS outperforms both supervised and existing state-of-the-art self-supervised methods for all label fractions. Parenthesis next to the techniques represents the architecture used; for example, DINO(ViT B/16) represents ViT B/16 trained with DINO. In this table, we compare the F1 score on the test set. We observed that CASS outperformed the existing state-of-art self-supervised method using all label fractions and for both the architectures.}
\label{dermofitperformance}
\end{table*}

\begin{table*}[t]
\centering
\begin{tabular}{lllll}
\hline
\multicolumn{1}{c}{\multirow{2}{*}{Techniques}} & \multicolumn{1}{c}{\multirow{2}{*}{Backbone}} & \multicolumn{3}{l}{Testing F1 score} \\
\multicolumn{1}{c}{}                            & \multicolumn{1}{c}{}                           & 1\%       & 10\%       & 100\%       \\
\hline
DINO                                            & Resnet-50                                      &\textbf{0.63405±0.09}          &\textbf{0.92325±0.02819}            & 0.9900±0.0058          \\
CASS                                           & Resnet-50                                      & 0.40816±0.13          & 0.8925±0.0254          &\textbf{0.9909±
0.0032}
             \\
Transfer                                      & Resnet-50                                      &0.52±0.018          &0.9022±0.011            & 0.9899± 0.003            \\
\hline
DINO                                            & ViT B/16                                          &0.3211±0.071      &0.7529±0.044           &0.8841±
0.0052
           \\
CASS                                           & ViT B/16                                           & \textbf{0.3345±0.11}          & \textbf{0.7833±0.0259}           &\textbf{0.9279± 
0.0213}
             \\
Transfer                                      & ViT B/16                                           & 0.3017 ± 0.077         & 0.747±0.0245           & 0.8719± 0.017           \\
\hline
\end{tabular}
\caption{This table contains results on the brain tumor MRI classification dataset. While DINO outperformed CASS for 1\% and 10\% labeled training for CNN, CASS maintained its superiority for 100\% labeled training, albeit by just 0.09\%. Similarly, CASS outperformed DINO for all data regimes for Transformers, incrementally 1.34\% in for 1\%, 3.04\% for 10\%, and 4.38\% for 100\% labeled training. We observe that this margin is more significant than for biopsy images. Such results could be ascribed to the increase in dataset size and increasing learnable information.}
\label{brainMRIperformance}
\end{table*}

\begin{table*}[!h]
\centering
\begin{tabular}{lllll}
\hline
\multicolumn{1}{c}{\multirow{2}{*}{Techniques}} & \multicolumn{1}{c}{\multirow{2}{*}{Backbone}} & \multicolumn{3}{l}{Testing Balanced multi-class accuracy} \\
\multicolumn{1}{c}{}                            & \multicolumn{1}{c}{}                           & 1\%       & 10\%       & 100\%       \\
\hline
DINO                                            & Resnet-50                                      &0.328±0.0016         &0.3797±0.0027            &0.493±3.9e-05            \\
CASS                                           & Resnet-50                                      &\textbf{0.3617±0.0047}            &\textbf{0.41±0.0019}            &  \textbf{0.543±2.85e-05}           \\
Transfer                                      & Resnet-50                                      &0.2640±0.031           &0.3070±0.0121            &0.35±0.006           \\ 
\hline
DINO                                            & ViT B/16                                           & 0.3676± 0.012           &0.3998±0.056          &0.5408±0.001            \\
CASS                                           & ViT B/16                                           &\textbf{0.3973± 0.0465}           &\textbf{0.4395±0.0179}            &  \textbf{0.5819±0.0015}          \\
Transfer                                      & ViT B/16                                           &0.3074±0.0005           & 0.3586±0.0314           &  0.42±0.007  \\
\hline
\end{tabular}

\caption{Results for the ISIC-2019  dataset. Comparable to the official metrics used in the ISIC 2019 challenge and previous works
\cite{Tschandl2018TheHD, Gutman2018SkinLA}. 
The ISIC-2019 dataset is incredibly challenging due to class imbalance and because it comprises partially processed and inconsistent images with hard-to-classify classes. We use balanced multi-class accuracy as our metric, semantically equal to recall value. We observed that CASS consistently outperforms DINO by approximately 4\% for all label fractions with CNN and Transformer.}
\label{ISICperformance}
\end{table*}

\begin{figure*}[!h]
    \centering
    
    \includegraphics[width=0.4\linewidth]{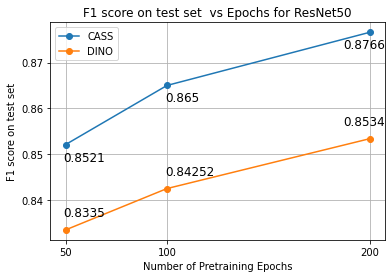}
    \hspace{1cm}
    \includegraphics[width=0.4\linewidth]{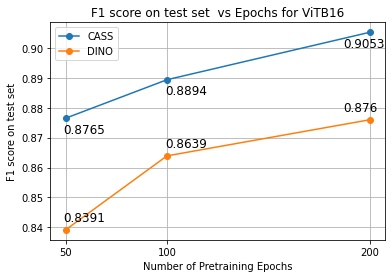}\\
    (a)
    

    \centering
    \centering  
    
    \includegraphics[width=0.4\linewidth]{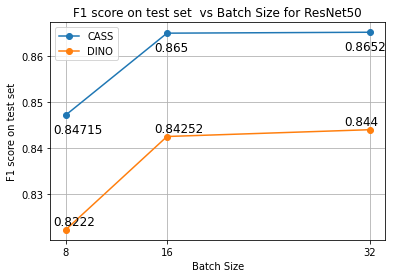}
    \hspace{1cm}
    \includegraphics[width=0.4\linewidth]{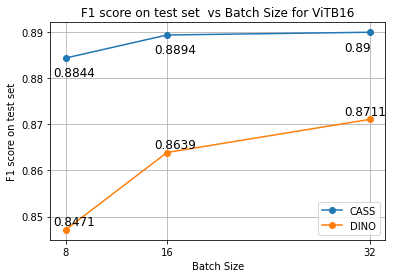}\\
    (b)
    
    \caption{In Figure a, we report the change in performance concerning the change in the number of pretraining epochs for DINO and CASS for ResNet-50 and ViTB/16, respectively. In Figure b, we report the change in performance concerning the change in the number of pretraining batch sizes for DINO and CASS for ResNet-50 and ViTB/16, respectively. These ablation studies were conducted on the autoimmune dataset while keeping the other hyper-parameters the same during pretraining and downstream finetuning.}
\label{fig:batchsize}
\end{figure*}

\subsection{Ablation Studies}

As mentioned in Section \ref{ssl-med}, existing self-supervised methods experience a drop in classification performance when trained for a reduced number of pretraining epochs and batch size. We performed ablation studies to study the effect of change in performance for ResNet-50 and ViTB/16 pre-trained with CASS and DINO on the autoimmune dataset. Additional ablation studies have been provided in Appendix \ref{aas-bs}. 
\subsubsection{Change in Pretraining Epochs and Batch Size}
\label{epoch-var}

This section compares the performance change in CASS and DINO when pretrained by varying the number of epochs or batch size and then E2E finetuned with 100\% labels over the autoimmune dataset. We compared the mean-variance over CNN and Transformer trained with the two techniques to study the robustness and reported these results in Figure \ref{fig:batchsize}. 

\paragraph{Change in Pretraining Epochs} The recorded mean-variance in performance for ResNet-50 and ViTB-16 trained with CASS and DINO with change in the number of pretraining epochs is $1.79\times10^{-4}$ and $2.265\times10^{-4}$, respectively. Based on these results, we observed that CASS-trained models have less variance, i.e., they are more robust to change in the number of pretraining epochs. 

\paragraph{Change in Pretraining Epochs} We studied the change in performance for batch sizes 8, 16, and 32 on the autoimmune dataset with CASS and DINO. We observe that the mean-variance in performance for ResNet-50 and ViTB-16 trained with CASS and DINO with change in batch size for CASS and DINO is $0.584\times10^{-4}$ and $1.5\times10^{-4}$, respectively. Hence, CASS is much more robust to changes in pretraining batch size than DINO.


\subsection{Comparison with BYOL}
\label{BYOL-cmp}
In addition to comparison with DINO, we compare with an additional contrastive technique Bootstrap your own Latent - BYOL \cite{grill2020bootstrap} in this section, and a reconstructive self-supervised technique Masked Auto-encoders or MAE \cite{he2022masked} in Appendix \ref{mae-cmp-appendix}. For these experiments, we pre-trained the models for 100 epochs followed by 50 epochs of E2E finetuning with 100\% using label fraction on the autoimmune and the dermofit dataset.\footnote{We attempted to broaden the scope of BYOL by incorporating ViT-Base/16 since BYOL was initially only implemented for CNNs. However, we were unable to fit the model on an RTX8000. We tried to reduce the batch size and implement gradient accumulation to accommodate BYOL with ViT on a single GPU. However, even with a batch size of two (as batch normalization in BYOL would not work with a smaller batch size) and a gradient accumulation step of eight, we were still unable to fit the model on a single RTX8000. Therefore, we present the corresponding results on ResNet50 in Table \ref{byol-comparison-table}.}

\begin{table}[!htb]
\centering
\begin{tabular}{llll}
\hline
Dataset                                                                        & Pretraining & Test F1 Score          & Architecture \\
\hline
\multirow{3}{*}{\begin{tabular}[c]{@{}l@{}}Autoimmune \\ Dataset\end{tabular}} & BYOL        & 0.8332±0.0015         & ResNet-50 \\
                                                                               & CASS        & \textbf{0.8650±0.0001} & ResNet-50\\
                                                                               & DINO        & 0.84252±0.008          & ResNet-50\\
                                                                               \hline
\multirow{3}{*}{\begin{tabular}[c]{@{}l@{}}Dermofit \\ Dataset\end{tabular}}   & BYOL        & 0.6642±0.0026          & ResNet-50\\
                                                                               & CASS        & \textbf{0.7132±0.0003} & ResNet-50\\
                                                                               & DINO        &0.6775±0.0005& ResNet-50\\\hline         
\end{tabular}
\caption{Comparison of BYOL, CASS, and DINO on the Autoimmune and the Dermofit dataset using ResNet-50. We observed that CASS outperforms BYOL by about 2.18\% on the Autoimmune dataset and by around 5\% on the Dermofit dataset. We also observed that in both cases BYOL performed worse than both CASS and DINO.}
\label{byol-comparison-table}
\end{table}

\subsection{Comparison with reconstructive self-supervised method - Masked Autoencoders (MAE)}

\label{mae-cmp-appendix}

In this section, we compare the performance of CASS with Masked Autoencoders (MAE) \cite{he2022masked}. We train CASS, DINO, and MAE on the autoimmune and dermofit datasets for this experimentation. Since MAE are only available for Vision Transformers, we compare ViT Base/16 trained with CASS and MAE. For this experiment, we followed the standard pre-training procedure for 100 epochs and fine-tuning for 50 epochs. We present the corresponding results for the autoimmune disease dataset and for the dermofit dataset in Table \ref{mae-comparison}.

\begin{table}[!htb]
\centering
\begin{tabular}{llll}
\hline
Dataset                                                                        & Pretraining & Test F1 Score         & Architecture \\
\hline
\multirow{3}{*}{\begin{tabular}[c]{@{}l@{}}Autoimmune \\ Dataset\end{tabular}} & DINO        & 0.8639± 0.002          & ViT B-16\\
                                                                               & CASS        & \textbf{0.8894±0.005} & ViT B-16\\
                                                                    
                                                                          &     MAE & 0.8369±0.0053& ViT B-16\\
                                                                               \hline
\multirow{3}{*}{\begin{tabular}[c]{@{}l@{}}Dermofit \\ Dataset\end{tabular}}   
                                                                               & DINO        &0.6775±0.0005& ViT B-16\\
                                                                               & CASS        & \textbf{0.7132±0.0003}& ViT B-16 \\
                                                                               & MAE        & 0.5621±0.0049& ViT B-16 \\
                                                                               
                                                                                                                                 \hline         
\end{tabular}

\caption{We observed that CASS performs 8.46\% better than MAE on the autoimmune dataset. Similarly, for performance comparison for CASS and MAE on the Dermofit dataset trained from scratch for 100 epochs followed by 50 epochs of end-to-end finetuning on the dermofit dataset. We observed that CASS outperforms MAE  by around 10\% in this scenario. }
\label{mae-comparison}
\end{table}


   

\section{Conclusion}

Based on our experimentation with four diverse medical imaging datasets, we conclude quantitatively that CASS outperforms the state-of-the-art self-supervised methods in most cases for data-efficient training for CNNs and in all cases for Vision Transformers. On average, CASS improves classification performance over DINO by 3.8\% with 1\% labeled data, 5.9\% with 10\% labeled data, and 10.13\% with 100\% labeled data, and is trained in 69\% less time. Moreover, CASS is more robust to changes in batch size and pretraining epochs. Therefore, we can conclude that CASS is computationally efficient for medical image analysis, performs better, and overcomes some of the shortcomings of existing self-supervised techniques. CASS, with its increased ease of accessibility and better performance, will catalyze medical imaging research to help improve healthcare solutions and propagate these advancements in state-of-the-art techniques to deep practical learning in developing countries and practitioners with limited resources to develop new solutions for underrepresented and emerging diseases.

\paragraph{Impact of CASS in the context of healthcare}
 Additionally, with CASS, researchers can begin medical image analysis, even with a small amount of the overall dataset or even if only a small portion is labeled. This would be highly helpful for emerging and rare diseases. Reduced reliance on intensive computational resources means large institutions can reduce computation budgets. Simultaneously, small institutions/practitioners with limited computing power can also run this state-of-the-art model.

\paragraph{Limitations} 
In this study, we focused extensively on studying the effects and performance of our proposed method for small dataset sizes and in the context of limited computational resources. Furthermore, all the datasets used in our experimentation are restricted to academic and research use only. Although CASS performs better than existing self-supervised and supervised techniques, it is impossible to determine at inference time (without ground-truth labels) whether to pick the CNN or the Transformers arm of CASS.

\paragraph{Potential negative societal impact}
The autoimmune dataset is limited to a specific institution, and inferences drawn may not apply to other disease variants. Results depend on a set of markers and may require additional tests and medical history. Meta-data is not incorporated in CASS. Application in real-life scenarios requires clearance from health and safety governing bodies.

\paragraph{Acknowledgements}We would like to thank Prof. Elena Sizikova (Moore Sloan
Faculty Fellow, Center for Data Science (CDS), New York
University (NYU)) for her valuable feedback and NYU HPC
team for assisting us with our computational needs.

\newpage
\bibliography{mlhc-camera-ready-template/egbib}

\begin{thebibliography}{59}
\providecommand{\natexlab}[1]{#1}
\providecommand{\url}[1]{\texttt{#1}}
\expandafter\ifx\csname urlstyle\endcsname\relax
  \providecommand{\doi}[1]{doi: #1}\else
  \providecommand{\doi}{doi: \begingroup \urlstyle{rm}\Url}\fi

\bibitem[Amin et~al.(2022)Amin, Anjum, Sharif, Jabeen, Kadry, and
  Ger]{Amin2022ANM}
Javeria Amin, Muhammad~Almas Anjum, Muhammad Sharif, Saima Jabeen, Seifedine
  Kadry, and Pablo~Moreno Ger.
\newblock A new model for brain tumor detection using ensemble transfer
  learning and quantum variational classifier.
\newblock \emph{Computational Intelligence and Neuroscience}, 2022, 2022.

\bibitem[Assran et~al.(2022{\natexlab{a}})Assran, Balestriero, Duval, Bordes,
  Misra, Bojanowski, Vincent, Rabbat, and Ballas]{assran2022hidden}
Mahmoud Assran, Randall Balestriero, Quentin Duval, Florian Bordes, Ishan
  Misra, Piotr Bojanowski, Pascal Vincent, Michael Rabbat, and Nicolas Ballas.
\newblock The hidden uniform cluster prior in self-supervised learning.
\newblock \emph{arXiv preprint arXiv:2210.07277}, 2022{\natexlab{a}}.

\bibitem[Assran et~al.(2022{\natexlab{b}})Assran, Caron, Misra, Bojanowski,
  Bordes, Vincent, Joulin, Rabbat, and Ballas]{assran2022masked}
Mahmoud Assran, Mathilde Caron, Ishan Misra, Piotr Bojanowski, Florian Bordes,
  Pascal Vincent, Armand Joulin, Mike Rabbat, and Nicolas Ballas.
\newblock Masked siamese networks for label-efficient learning.
\newblock In \emph{Computer Vision--ECCV 2022: 17th European Conference, Tel
  Aviv, Israel, October 23--27, 2022, Proceedings, Part XXXI}, pages 456--473.
  Springer, 2022{\natexlab{b}}.

\bibitem[Azizi et~al.(2021{\natexlab{a}})Azizi, Mustafa, Ryan, Beaver, von
  Freyberg, Deaton, Loh, Karthikesalingam, Kornblith, Chen, Natarajan, and
  Norouzi]{Azizi2021BigSM}
Shekoofeh Azizi, Basil Mustafa, Fiona Ryan, Zach Beaver, Jana von Freyberg,
  Jonathan Deaton, Aaron Loh, Alan Karthikesalingam, Simon Kornblith, Ting
  Chen, Vivek Natarajan, and Mohammad Norouzi.
\newblock Big self-supervised models advance medical image classification.
\newblock \emph{2021 IEEE/CVF International Conference on Computer Vision
  (ICCV)}, pages 3458--3468, 2021{\natexlab{a}}.

\bibitem[Azizi et~al.(2021{\natexlab{b}})Azizi, Mustafa, Ryan, Beaver,
  Freyberg, Deaton, Loh, Karthikesalingam, Kornblith, Chen,
  et~al.]{azizi2021big}
Shekoofeh Azizi, Basil Mustafa, Fiona Ryan, Zachary Beaver, Jan Freyberg,
  Jonathan Deaton, Aaron Loh, Alan Karthikesalingam, Simon Kornblith, Ting
  Chen, et~al.
\newblock Big self-supervised models advance medical image classification.
\newblock In \emph{Proceedings of the IEEE/CVF International Conference on
  Computer Vision}, pages 3478--3488, 2021{\natexlab{b}}.

\bibitem[Bardes et~al.(2021)Bardes, Ponce, and LeCun]{bardes2021vicreg}
Adrien Bardes, Jean Ponce, and Yann LeCun.
\newblock Vicreg: Variance-invariance-covariance regularization for
  self-supervised learning.
\newblock \emph{arXiv preprint arXiv:2105.04906}, 2021.

\bibitem[Caron et~al.(2020{\natexlab{a}})Caron, Misra, Mairal, Goyal,
  Bojanowski, and Joulin]{Caron2020UnsupervisedLO}
Mathilde Caron, Ishan Misra, Julien Mairal, Priya Goyal, Piotr Bojanowski, and
  Armand Joulin.
\newblock Unsupervised learning of visual features by contrasting cluster
  assignments.
\newblock \emph{ArXiv}, abs/2006.09882, 2020{\natexlab{a}}.

\bibitem[Caron et~al.(2020{\natexlab{b}})Caron, Misra, Mairal, Goyal,
  Bojanowski, and Joulin]{caron2020unsupervised}
Mathilde Caron, Ishan Misra, Julien Mairal, Priya Goyal, Piotr Bojanowski, and
  Armand Joulin.
\newblock Unsupervised learning of visual features by contrasting cluster
  assignments.
\newblock \emph{Advances in Neural Information Processing Systems},
  33:\penalty0 9912--9924, 2020{\natexlab{b}}.

\bibitem[Caron et~al.(2021)Caron, Touvron, Misra, J\'egou, Mairal, Bojanowski,
  and Joulin]{caron2021emerging}
Mathilde Caron, Hugo Touvron, Ishan Misra, Herv\'e J\'egou, Julien Mairal,
  Piotr Bojanowski, and Armand Joulin.
\newblock Emerging properties in self-supervised vision transformers.
\newblock In \emph{Proceedings of the International Conference on Computer
  Vision (ICCV)}, 2021.

\bibitem[Cassidy et~al.(2022)Cassidy, Kendrick, Brodzicki, Jaworek-Korjakowska,
  and Yap]{cassidy2022analysis}
Bill Cassidy, Connah Kendrick, Andrzej Brodzicki, Joanna Jaworek-Korjakowska,
  and Moi~Hoon Yap.
\newblock Analysis of the isic image datasets: usage, benchmarks and
  recommendations.
\newblock \emph{Medical Image Analysis}, 75:\penalty0 102305, 2022.

\bibitem[Chen et~al.(2020{\natexlab{a}})Chen, Kornblith, Norouzi, and
  Hinton]{chen2020simple}
Ting Chen, Simon Kornblith, Mohammad Norouzi, and Geoffrey Hinton.
\newblock A simple framework for contrastive learning of visual
  representations.
\newblock \emph{arXiv preprint arXiv:2002.05709}, 2020{\natexlab{a}}.

\bibitem[Chen et~al.(2020{\natexlab{b}})Chen, Kornblith, Norouzi, and
  Hinton]{Chen2020ASF}
Ting Chen, Simon Kornblith, Mohammad Norouzi, and Geoffrey~E. Hinton.
\newblock A simple framework for contrastive learning of visual
  representations.
\newblock \emph{ArXiv}, abs/2002.05709, 2020{\natexlab{b}}.

\bibitem[Chen and He(2021)]{Chen2021ExploringSS}
Xinlei Chen and Kaiming He.
\newblock Exploring simple siamese representation learning.
\newblock \emph{2021 IEEE/CVF Conference on Computer Vision and Pattern
  Recognition (CVPR)}, pages 15745--15753, 2021.

\bibitem[Cheng(2017)]{Cheng2017}
Jun Cheng.
\newblock {brain tumor dataset}.
\newblock 4 2017.
\newblock \doi{10.6084/m9.figshare.1512427.v5}.
\newblock URL
  \url{https://figshare.com/articles/dataset/brain_tumor_dataset/1512427}.

\bibitem[Combalia et~al.(2019)Combalia, Codella, Rotemberg, Helba, Vilaplana,
  Reiter, Halpern, Puig, and Malvehy]{Combalia2019BCN20000DL}
Marc Combalia, Noel C.~F. Codella, Veronica~M Rotemberg, Brian Helba,
  Ver{\'o}nica Vilaplana, Ofer Reiter, Allan~C. Halpern, Susana Puig, and Josep
  Malvehy.
\newblock Bcn20000: Dermoscopic lesions in the wild.
\newblock \emph{ArXiv}, abs/1908.02288, 2019.

\bibitem[d'Ascoli et~al.(2021)d'Ascoli, Touvron, Leavitt, Morcos, Biroli, and
  Sagun]{d2021convit}
St{\'e}phane d'Ascoli, Hugo Touvron, Matthew Leavitt, Ari Morcos, Giulio
  Biroli, and Levent Sagun.
\newblock Convit: Improving vision transformers with soft convolutional
  inductive biases.
\newblock \emph{arXiv preprint arXiv:2103.10697}, 2021.

\bibitem[Deng et~al.(2009)Deng, Dong, Socher, Li, Li, and
  Fei-Fei]{deng2009imagenet}
Jia Deng, Wei Dong, Richard Socher, Li-Jia Li, Kai Li, and Li~Fei-Fei.
\newblock Imagenet: A large-scale hierarchical image database.
\newblock In \emph{2009 IEEE conference on computer vision and pattern
  recognition}, pages 248--255. Ieee, 2009.

\bibitem[Dosovitskiy et~al.(2020)Dosovitskiy, Beyer, Kolesnikov, Weissenborn,
  Zhai, Unterthiner, Dehghani, Minderer, Heigold, Gelly,
  et~al.]{dosovitskiy2020image}
Alexey Dosovitskiy, Lucas Beyer, Alexander Kolesnikov, Dirk Weissenborn,
  Xiaohua Zhai, Thomas Unterthiner, Mostafa Dehghani, Matthias Minderer, Georg
  Heigold, Sylvain Gelly, et~al.
\newblock An image is worth 16x16 words: Transformers for image recognition at
  scale.
\newblock \emph{arXiv preprint arXiv:2010.11929}, 2020.

\bibitem[Ehrenfeld et~al.(2020)Ehrenfeld, Tincani, Andreoli, Cattalini,
  Greenbaum, Kanduc, Alijotas-Reig, Zinserling, Semenova, Amital,
  et~al.]{ehrenfeld2020covid}
Michael Ehrenfeld, Angela Tincani, Laura Andreoli, Marco Cattalini, Assaf
  Greenbaum, Darja Kanduc, Jaume Alijotas-Reig, Vsevolod Zinserling, Natalia
  Semenova, Howard Amital, et~al.
\newblock Covid-19 and autoimmunity.
\newblock \emph{Autoimmunity reviews}, 19\penalty0 (8):\penalty0 102597, 2020.

\bibitem[Fisher and Rees(2017)]{Dermofit}
Robert Fisher and Jonathan Rees.
\newblock Dermofit project datasets.
\newblock 2017.
\newblock URL \url{https://homepages.inf.ed.ac.uk/rbf/DERMOFIT/datasets.htm}.

\bibitem[Galeotti and Bayry(2020)]{galeotti2020autoimmune}
Caroline Galeotti and Jagadeesh Bayry.
\newblock Autoimmune and inflammatory diseases following covid-19.
\newblock \emph{Nature Reviews Rheumatology}, 16\penalty0 (8):\penalty0
  413--414, 2020.

\bibitem[Gessert et~al.(2020)Gessert, Nielsen, Shaikh, Werner, and
  Schlaefer]{Gessert2020SkinLC}
Nils Gessert, Maximilian Nielsen, Mohsin Shaikh, Ren{\'e} Werner, and
  A.~Schlaefer.
\newblock Skin lesion classification using ensembles of multi-resolution
  efficientnets with meta data.
\newblock \emph{MethodsX}, 7, 2020.

\bibitem[Ghesu et~al.(2022)Ghesu, Georgescu, Mansoor, Yoo, Neumann, Patel,
  Vishwanath, Balter, Cao, Grbic, et~al.]{ghesu2022self}
Florin~C Ghesu, Bogdan Georgescu, Awais Mansoor, Youngjin Yoo, Dominik Neumann,
  Pragneshkumar Patel, RS~Vishwanath, James~M Balter, Yue Cao, Sasa Grbic,
  et~al.
\newblock Self-supervised learning from 100 million medical images.
\newblock \emph{arXiv preprint arXiv:2201.01283}, 2022.

\bibitem[Gong et~al.(2022)Gong, Khurana, Rouditchenko, and
  Glass]{Gong2022CMKDCC}
Yuan Gong, Sameer Khurana, Andrew Rouditchenko, and James~R. Glass.
\newblock Cmkd: Cnn/transformer-based cross-model knowledge distillation for
  audio classification.
\newblock \emph{ArXiv}, abs/2203.06760, 2022.

\bibitem[Gou et~al.(2021)Gou, Yu, Maybank, and Tao]{gou2021knowledge}
Jianping Gou, Baosheng Yu, Stephen~J Maybank, and Dacheng Tao.
\newblock Knowledge distillation: A survey.
\newblock \emph{International Journal of Computer Vision}, 129\penalty0
  (6):\penalty0 1789--1819, 2021.

\bibitem[Grill et~al.(2020{\natexlab{a}})Grill, Strub, Altch{\'e}, Tallec,
  Richemond, Buchatskaya, Doersch, Avila~Pires, Guo, Gheshlaghi~Azar,
  et~al.]{grill2020bootstrap}
Jean-Bastien Grill, Florian Strub, Florent Altch{\'e}, Corentin Tallec, Pierre
  Richemond, Elena Buchatskaya, Carl Doersch, Bernardo Avila~Pires, Zhaohan
  Guo, Mohammad Gheshlaghi~Azar, et~al.
\newblock Bootstrap your own latent-a new approach to self-supervised learning.
\newblock \emph{Advances in neural information processing systems},
  33:\penalty0 21271--21284, 2020{\natexlab{a}}.

\bibitem[Grill et~al.(2020{\natexlab{b}})Grill, Strub, Altch'e, Tallec,
  Richemond, Buchatskaya, Doersch, Pires, Guo, Azar, Piot, Kavukcuoglu, Munos,
  and Valko]{Grill2020BootstrapYO}
Jean-Bastien Grill, Florian Strub, Florent Altch'e, Corentin Tallec, Pierre~H.
  Richemond, Elena Buchatskaya, Carl Doersch, Bernardo~{\'A}vila Pires,
  Zhaohan~Daniel Guo, Mohammad~Gheshlaghi Azar, Bilal Piot, Koray Kavukcuoglu,
  R{\'e}mi Munos, and Michal Valko.
\newblock Bootstrap your own latent: A new approach to self-supervised
  learning.
\newblock \emph{ArXiv}, abs/2006.07733, 2020{\natexlab{b}}.

\bibitem[Guo et~al.(2022)Guo, Xiong, Zhong, Wang, Guo, Han, and
  Huang]{guo2022cross}
Sheng Guo, Zihua Xiong, Yujie Zhong, Limin Wang, Xiaobo Guo, Bing Han, and
  Weilin Huang.
\newblock Cross-architecture self-supervised video representation learning.
\newblock In \emph{Proceedings of the IEEE/CVF Conference on Computer Vision
  and Pattern Recognition}, pages 19270--19279, 2022.

\bibitem[Gutman et~al.(2018)Gutman, Codella, Celebi, Helba, Marchetti, Mishra,
  and Halpern]{Gutman2018SkinLA}
David~A. Gutman, Noel C.~F. Codella, M.~E. Celebi, Brian Helba, Michael~Armando
  Marchetti, Nabin~K. Mishra, and Allan~C. Halpern.
\newblock Skin lesion analysis toward melanoma detection: A challenge at the
  2017 international symposium on biomedical imaging (isbi), hosted by the
  international skin imaging collaboration (isic).
\newblock \emph{2018 IEEE 15th International Symposium on Biomedical Imaging
  (ISBI 2018)}, pages 168--172, 2018.

\bibitem[He et~al.(2016)He, Zhang, Ren, and Sun]{He2016DeepRL}
Kaiming He, X.~Zhang, Shaoqing Ren, and Jian Sun.
\newblock Deep residual learning for image recognition.
\newblock \emph{2016 IEEE Conference on Computer Vision and Pattern Recognition
  (CVPR)}, pages 770--778, 2016.

\bibitem[He et~al.(2020)He, Fan, Wu, Xie, and Girshick]{He2020MomentumCF}
Kaiming He, Haoqi Fan, Yuxin Wu, Saining Xie, and Ross~B. Girshick.
\newblock Momentum contrast for unsupervised visual representation learning.
\newblock \emph{2020 IEEE/CVF Conference on Computer Vision and Pattern
  Recognition (CVPR)}, pages 9726--9735, 2020.

\bibitem[He et~al.(2022)He, Chen, Xie, Li, Doll{\'a}r, and
  Girshick]{he2022masked}
Kaiming He, Xinlei Chen, Saining Xie, Yanghao Li, Piotr Doll{\'a}r, and Ross
  Girshick.
\newblock Masked autoencoders are scalable vision learners.
\newblock In \emph{Proceedings of the IEEE/CVF Conference on Computer Vision
  and Pattern Recognition}, pages 16000--16009, 2022.

\bibitem[Izmailov et~al.(2018)Izmailov, Podoprikhin, Garipov, Vetrov, and
  Wilson]{Izmailov2018AveragingWL}
Pavel Izmailov, Dmitrii Podoprikhin, T.~Garipov, Dmitry~P. Vetrov, and
  Andrew~Gordon Wilson.
\newblock Averaging weights leads to wider optima and better generalization.
\newblock \emph{ArXiv}, abs/1803.05407, 2018.

\bibitem[Kang et~al.(2021)Kang, Ullah, and Gwak]{s21062222}
Jaeyong Kang, Zahid Ullah, and Jeonghwan Gwak.
\newblock Mri-based brain tumor classification using ensemble of deep features
  and machine learning classifiers.
\newblock \emph{Sensors}, 21\penalty0 (6), 2021.
\newblock ISSN 1424-8220.
\newblock \doi{10.3390/s21062222}.
\newblock URL \url{https://www.mdpi.com/1424-8220/21/6/2222}.

\bibitem[Khan et~al.(2020)Khan, Sohail, Zahoora, and Qureshi]{khan2020survey}
Asifullah Khan, Anabia Sohail, Umme Zahoora, and Aqsa~Saeed Qureshi.
\newblock A survey of the recent architectures of deep convolutional neural
  networks.
\newblock \emph{Artificial intelligence review}, 53\penalty0 (8):\penalty0
  5455--5516, 2020.

\bibitem[Lerner et~al.(2015)Lerner, Jeremias, and Matthias]{lerner2015world}
Aaron Lerner, Patricia Jeremias, and Torsten Matthias.
\newblock The world incidence and prevalence of autoimmune diseases is
  increasing.
\newblock \emph{Int J Celiac Dis}, 3\penalty0 (4):\penalty0 151--5, 2015.

\bibitem[Li et~al.(2021)Li, Tang, Wang, Peng, Wang, Liang, and
  Chang]{li2021bossnas}
Changlin Li, Tao Tang, Guangrun Wang, Jiefeng Peng, Bing Wang, Xiaodan Liang,
  and Xiaojun Chang.
\newblock Bossnas: Exploring hybrid cnn-transformers with block-wisely
  self-supervised neural architecture search.
\newblock In \emph{Proceedings of the IEEE/CVF International Conference on
  Computer Vision}, pages 12281--12291, 2021.

\bibitem[Lin et~al.(2017)Lin, Goyal, Girshick, He, and
  Doll{\'a}r]{Lin2017FocalLF}
Tsung-Yi Lin, Priya Goyal, Ross~B. Girshick, Kaiming He, and Piotr Doll{\'a}r.
\newblock Focal loss for dense object detection.
\newblock \emph{2017 IEEE International Conference on Computer Vision (ICCV)},
  pages 2999--3007, 2017.

\bibitem[Liu et~al.(2020)Liu, Sawalha, and Lu]{Liu2020COVID19AA}
Yu~Liu, Amr~H. Sawalha, and Qianjin Lu.
\newblock Covid-19 and autoimmune diseases.
\newblock \emph{Current Opinion in Rheumatology}, 33:\penalty0 155 -- 162,
  2020.

\bibitem[Liu et~al.(2021{\natexlab{a}})Liu, Lin, Cao, Hu, Wei, Zhang, Lin, and
  Guo]{Liu_2021_ICCV}
Ze~Liu, Yutong Lin, Yue Cao, Han Hu, Yixuan Wei, Zheng Zhang, Stephen Lin, and
  Baining Guo.
\newblock Swin transformer: Hierarchical vision transformer using shifted
  windows.
\newblock In \emph{Proceedings of the IEEE/CVF International Conference on
  Computer Vision (ICCV)}, pages 10012--10022, October 2021{\natexlab{a}}.

\bibitem[Liu et~al.(2021{\natexlab{b}})Liu, Lin, Cao, Hu, Wei, Zhang, Lin, and
  Guo]{liu2021Swin}
Ze~Liu, Yutong Lin, Yue Cao, Han Hu, Yixuan Wei, Zheng Zhang, Stephen Lin, and
  Baining Guo.
\newblock Swin transformer: Hierarchical vision transformer using shifted
  windows.
\newblock In \emph{Proceedings of the IEEE/CVF International Conference on
  Computer Vision (ICCV)}, 2021{\natexlab{b}}.

\bibitem[Liu et~al.(2022{\natexlab{a}})Liu, Hu, Lin, Yao, Xie, Wei, Ning, Cao,
  Zhang, Dong, Wei, and Guo]{liu2021swinv2}
Ze~Liu, Han Hu, Yutong Lin, Zhuliang Yao, Zhenda Xie, Yixuan Wei, Jia Ning, Yue
  Cao, Zheng Zhang, Li~Dong, Furu Wei, and Baining Guo.
\newblock Swin transformer v2: Scaling up capacity and resolution.
\newblock In \emph{International Conference on Computer Vision and Pattern
  Recognition (CVPR)}, 2022{\natexlab{a}}.

\bibitem[Liu et~al.(2022{\natexlab{b}})Liu, Mao, Wu, Feichtenhofer, Darrell,
  and Xie]{liu2022convnet}
Zhuang Liu, Hanzi Mao, Chao-Yuan Wu, Christoph Feichtenhofer, Trevor Darrell,
  and Saining Xie.
\newblock A convnet for the 2020s.
\newblock \emph{Proceedings of the IEEE/CVF Conference on Computer Vision and
  Pattern Recognition (CVPR)}, 2022{\natexlab{b}}.

\bibitem[Matsoukas et~al.(2021)Matsoukas, Haslum, Soderberg, and
  Smith]{Matsoukas2021IsIT}
Christos Matsoukas, Johan~Fredin Haslum, Magnus~P Soderberg, and Kevin Smith.
\newblock Is it time to replace cnns with transformers for medical images?
\newblock \emph{ArXiv}, abs/2108.09038, 2021.

\bibitem[Nickparvar(2021)]{msoud_nickparvar_2021}
Msoud Nickparvar.
\newblock Brain tumor mri dataset, 2021.
\newblock URL \url{https://www.kaggle.com/dsv/2645886}.

\bibitem[Picard(2021)]{picard2021torch}
David Picard.
\newblock Torch. manual\_seed (3407) is all you need: On the influence of
  random seeds in deep learning architectures for computer vision.
\newblock \emph{arXiv preprint arXiv:2109.08203}, 2021.

\bibitem[Raghu et~al.(2019)Raghu, Zhang, Kleinberg, and
  Bengio]{raghu2019transfusion}
Maithra Raghu, Chiyuan Zhang, Jon Kleinberg, and Samy Bengio.
\newblock Transfusion: Understanding transfer learning for medical imaging.
\newblock \emph{Advances in neural information processing systems}, 32, 2019.

\bibitem[Raghu et~al.(2021)Raghu, Unterthiner, Kornblith, Zhang, and
  Dosovitskiy]{Raghu2021DoVT}
Maithra Raghu, Thomas Unterthiner, Simon Kornblith, Chiyuan Zhang, and Alexey
  Dosovitskiy.
\newblock Do vision transformers see like convolutional neural networks?
\newblock In \emph{NeurIPS}, 2021.

\bibitem[Ronneberger et~al.(2015)Ronneberger, Fischer, and
  Brox]{ronneberger2015u}
Olaf Ronneberger, Philipp Fischer, and Thomas Brox.
\newblock U-net: Convolutional networks for biomedical image segmentation.
\newblock In \emph{International Conference on Medical image computing and
  computer-assisted intervention}, pages 234--241. Springer, 2015.

\bibitem[Singh and Cirrone(2023)]{singh2023data}
Pranav Singh and Jacopo Cirrone.
\newblock A data-efficient deep learning framework for segmentation and
  classification of histopathology images.
\newblock In \emph{Computer Vision--ECCV 2022 Workshops: Tel Aviv, Israel,
  October 23--27, 2022, Proceedings, Part III}, pages 385--405. Springer, 2023.

\bibitem[Sriram et~al.(2021)Sriram, Muckley, Sinha, Shamout, Pineau, Geras,
  Azour, Aphinyanaphongs, Yakubova, and Moore]{sriram2021covid}
Anuroop Sriram, Matthew Muckley, Koustuv Sinha, Farah Shamout, Joelle Pineau,
  Krzysztof~J Geras, Lea Azour, Yindalon Aphinyanaphongs, Nafissa Yakubova, and
  William Moore.
\newblock Covid-19 prognosis via self-supervised representation learning and
  multi-image prediction.
\newblock \emph{arXiv preprint arXiv:2101.04909}, 2021.

\bibitem[Stafford et~al.(2020)Stafford, Kellermann, Mossotto, Beattie,
  MacArthur, and Ennis]{Stafford2020ASR}
I.~S. Stafford, M~Kellermann, E~Mossotto, Robert~Mark Beattie, Ben~D.
  MacArthur, and Sarah Ennis.
\newblock A systematic review of the applications of artificial intelligence
  and machine learning in autoimmune diseases.
\newblock \emph{NPJ Digital Medicine}, 3, 2020.

\bibitem[Touvron et~al.(2020)Touvron, Cord, Douze, Massa, Sablayrolles, and
  J{\'e}gou]{touvron2020training}
Hugo Touvron, Matthieu Cord, Matthijs Douze, Francisco Massa, Alexandre
  Sablayrolles, and Herv{\'e} J{\'e}gou.
\newblock Training data-efficient image transformers \& distillation through
  attention. arxiv 2020.
\newblock \emph{arXiv preprint arXiv:2012.12877}, 2020.

\bibitem[Touvron et~al.(2021)Touvron, Cord, Douze, Massa, Sablayrolles, and
  Jegou]{pmlr-v139-touvron21a}
Hugo Touvron, Matthieu Cord, Matthijs Douze, Francisco Massa, Alexandre
  Sablayrolles, and Herve Jegou.
\newblock Training data-efficient image transformers \& amp; distillation
  through attention.
\newblock In \emph{International Conference on Machine Learning}, volume 139,
  pages 10347--10357, July 2021.

\bibitem[Tsakalidou et~al.(2022)Tsakalidou, Mitsou, and
  Papakostas]{tsakalidou2022computer}
Viktoria~N Tsakalidou, Pavlina Mitsou, and George~A Papakostas.
\newblock Computer vision in autoimmune diseases diagnosis—current status and
  perspectives.
\newblock In \emph{Computational Vision and Bio-Inspired Computing}, pages
  571--586. Springer, 2022.

\bibitem[Tschandl et~al.(2018)Tschandl, Rosendahl, and
  Kittler]{Tschandl2018TheHD}
Philipp Tschandl, Cliff Rosendahl, and Harald Kittler.
\newblock The ham10000 dataset, a large collection of multi-source
  dermatoscopic images of common pigmented skin lesions.
\newblock \emph{Scientific Data}, 5, 2018.

\bibitem[{Van Buren} et~al.(2022){Van Buren}, Li, Zhong, Ding, Puranik, Loomis,
  Razavian, and Niewold]{VANBUREN2022113233}
Kayla {Van Buren}, Yi~Li, Fanghao Zhong, Yuan Ding, Amrutesh Puranik,
  Cynthia~A. Loomis, Narges Razavian, and Timothy~B. Niewold.
\newblock Artificial intelligence and deep learning to map immune cell types in
  inflamed human tissue.
\newblock \emph{Journal of Immunological Methods}, 505:\penalty0 113233, 2022.
\newblock ISSN 0022-1759.
\newblock \doi{https://doi.org/10.1016/j.jim.2022.113233}.
\newblock URL
  \url{https://www.sciencedirect.com/science/article/pii/S0022175922000205}.

\bibitem[Wightman(2019)]{rw2019timm}
Ross Wightman.
\newblock Pytorch image models.
\newblock \url{https://github.com/rwightman/pytorch-image-models}, 2019.

\bibitem[Yadav and Jadhav(2019)]{yadav2019deep}
Samir~S Yadav and Shivajirao~M Jadhav.
\newblock Deep convolutional neural network based medical image classification
  for disease diagnosis.
\newblock \emph{Journal of Big Data}, 6\penalty0 (1):\penalty0 1--18, 2019.

\end{thebibliography}

\appendix

\section{CASS Pretraining Algorithm}
The core self-supervised algorithm used to train CASS with a CNN (R) and a Transformer (T) is described in Algorithm \ref{algorithm}. Here, for each sample for the data loader, we perform steps 1 through 4. When repeated once over the entire dataset, this would be one epoch of self-supervised pretraining. Multiple repetitions of this training over the entire dataset would yield multiple epochs of self-supervised training. The loss used is described in Equation \ref{loss_eq}. Finally, after pretraining, we save the CNN and Transformer for downstream finetuning.

\algrenewcommand\algorithmicrequire{\textbf{Input:}}
\algrenewcommand\algorithmicensure{\textbf{Output:}}

\begin{algorithm}[tb]
   \caption{CASS self-supervised Pretraining algorithm}
   \begin{algorithmic}[1]
  \Require Unlabeled same augmented images from the training set $x'$
\State $R = cnn( x')$ \Comment{taking logits output from CNN}
\State $T = vit( x')$ \Comment{taking logits output from ViT}
\State Calculate loss using Equation~\ref{loss_eq}.
\State Backpropogate loss.
\end{algorithmic}
\label{algorithm}
\end{algorithm}

\section{Additional Ablation Studies}
\label{aas-bs}
\subsection{Batch size}

We studied the effect of change in batch size on the autoimmune dataset in Section \ref{epoch-var}. Similarly, in this section, we study the effect of varying the batch size on the brain MRI classification dataset. In the standard implementation of CASS, we used a batch size of 16; here, we showed results for batch sizes 8 and 32. The largest batch size we could run was 34 on a single GPU of 48 GB video memory. Hence 32 was the biggest batch size we showed in our results. We present these results in Table \ref{batchsize_bmri}. Similar to the results in Section \ref{epoch-var}, performance decreases as we reduce the batch size and increases slightly as we increase the batch size for both CNN and Transformer. 

\begin{table}[!htb]
\centering
\begin{tabular}{lll}
\hline
Batch Size & CNN F1 Score                       & Transformer F1 Score \\
\hline
8          & 0.9895±0.0025                      & 0.9198±0.0109       \\
16         & \multicolumn{1}{c}{0.9909± 0.0032} & 0.9279± 0.0213       \\
32         & 0.991±0.011                       & 0.9316±0.006        \\
\hline
\end{tabular}
\caption{This table represents the results for different batch sizes on the brain MRI classification dataset.  We maintain the downstream batch size constant in all three cases, following the standard experimental setup mentioned in Appendix \ref{ssl-train} and \ref{sup-train}. These results are on the test set after E2E fine-tuning with 100\% labels.}
\label{batchsize_bmri}
\end{table}

\subsection{Change in pretraining epochs}

As standard, we pretrained CASS for 100 epochs in all cases. However, existing self-supervised techniques are plagued with a marked loss in performance with a reduction in pretraining epochs. We reported results in Section \ref{epoch-var} to study this effect of CASS and DINO on the autoimmune dataset. Additionally, in this section, we report results for pretraining CASS for 300 epochs on the autoimmune and brain tumor MRI datasets. We reported these results in Table \ref{epochs_atm} and \ref{epochs_bmri}, respectively. We observed a slight gain in performance when we increased the epochs from 100 to 200 but minimal gain beyond that. We also studied the effect of longer pretraining on the  brain tumor MRI classification dataset and presented these results in Table \ref{epochs_bmri}. 
\begin{table}[!htb]
\centering
\begin{tabular}{lll}
\hline
\multicolumn{1}{c}{Epochs} & CNN F1 Score  & Transformer F1 Score \\
\hline
50                         & 0.8521±0.0007 & 0.8765± 0.0021       \\
100                        & 0.8650±0.0001 & 0.8894±0.005         \\
200                        & 0.8766±0.001  & 0.9053±0.008         \\
300                        & 0.8777±0.004  & 0.9091±8.2e-5       \\
\hline
\end{tabular}
\caption{ Performance comparison over a varied number of epochs on the brain tumor MRI classification dataset, from 50 to 300 epochs, the downstream training procedure, and the CNN-Transformer combination is kept constant across all the four experiments, only the number of self-supervised pretraining epochs were changed.}
\label{epochs_atm}
\end{table}

\begin{table}[!htb]
\centering
\begin{tabular}{lll}
\hline
Epochs & CNN F1 Score                       & Transformer F1 Score \\
\hline
50                              & 0.9795±0.0109                      & 0.9262±0.0181       \\
100                             & \multicolumn{1}{c}{0.9909± 0.0032} & 0.9279± 0.0213       \\
200                             & 0.9864±0.008                       & 0.9476±0.0012        \\
300                             & 0.9920±0.001                       & 0.9484±0.017   \\
\hline
\end{tabular}
\caption{ Performance comparison over a varied number of epochs, from 50 to 300 epochs, the downstream training procedure, and the CNN-transformer combination is kept constant across all four experiments; only the number of self-supervised epochs has been changed.}
\label{epochs_bmri}
\end{table}

\subsection{Augmentations}

Contrastive learning techniques are known to be highly dependent on augmentations. Recently, most self-supervised techniques have adopted BYOL \cite{Grill2020BootstrapYO}-like set of augmentations. DINO \cite{caron2021emerging} uses the same set of augmentations as BYOL, along with adding local-global cropping. We use a reduced set of BYOL augmentations for CASS and a few changes. For instance, we do not use solarize and Gaussian blur. Instead, we use affine transformations and random perspectives. This section studies the effect of adding BYOL-like augmentations to CASS. We report these results in Table \ref{augmentaions}. We observed that CASS-trained CNN is robust to changes in augmentations. On the other hand, the Transformer drops performance with changes in augmentations. A possible solution to regain this loss in performance for Transformer with a change in augmentation is using Gaussian blur, which converges the performance of CNN and the Transformer. 

\begin{table*}[!htb]
\label{table:augchnages}
\centering
\begin{tabular}{cll}
\hline
Augmentation Set                                    & CNN F1 Score    & Transformer F1 Score \\
\hline
CASS only                                           & 0.8650±0.0001   & 0.8894±0.005         \\
CASS + Solarize                                     & 0.8551±0.0004   & 0.81455±0.002        \\
CASS + Gaussian blur                                & 0.864±4.2e-05   & 0.8604±0.0029        \\

\multicolumn{1}{l}{CASS + Gaussian blur + Solarize} & 0.8573±2.59e-05 & 0.8513±0.0066       \\
\hline
\end{tabular}
\caption{We report the F1 metric of CASS trained with
a different set of augmentations for 100 epochs. While CASS-trained CNN fluctuates within a percent of its peak performance, CASS-trained Transformer drops performance with the addition of solarization and Gaussian blur. Interestingly, the two arms converged with the use of Gaussian blur.}
\label{augmentaions}
\end{table*}

\subsection{Optimization}
\label{optim-results}
In CASS, we use Adam optimizer for both CNN and Transformer. This is a shift from using SGD or stochastic gradient descent for CNNs. In this Table \ref{optimizer}, we report the performance of CASS-trained CNN and Transformer with the CNN using SGD and Adam optimizer. We observed that while the performance of CNN remained almost constant, the performance of the Transformer dropped by almost 6\% with CNN using SGD.

\begin{table*}[!htb]
\centering
\begin{tabular}{cll}
\hline
Optimiser for CNN                                   & CNN F1 Score    & Transformer F1 Score \\
\hline
Adam                                                & 0.8650±0.0001   & 0.8894±0.005         \\
SGD                                                 & 0.8648±0.0005   & 0.82355±0.0064       \\
\hline
\end{tabular}
\caption{We report the F1 metric of CASS trained with
a different set of optimizers for the CNN arm for 100 epochs. While there is no change in CNN's performance, the Transformer's performance drops around 6\% with SGD.}
\label{optimizer}
\end{table*}

\subsection{Using softmax and sigmoid layer in CASS}
\label{softmax-appendix}

As noted in Fig \ref{fig:CASS}, CASS doesn’t use a softmax layer like DINO \cite{caron2021emerging} before the computing loss. The output logits of the two networks have been used to combine the two architectures in a response-based knowledge distillation \cite{gou2021knowledge} manner instead of using soft labels from the softmax layer. In this section, we study the effect of using an additional softmax layer on CASS. Furthermore, we also study the effect of adding a sigmoid layer instead of a softmax layer and compare it with a CASS model that doesn’t use the sigmoid or the softmax layer. We present these results in Table \ref{softmax}. We observed that not using sigmoid and softmax layers in CASS yields the best result for both CNN and Transformers.

\begin{table*}[!htb]
\centering
\begin{tabular}{lll}
\hline
Techniques & CNN F1 Score  & Transformer F1 Score \\
\hline
Without Sigmoid or Softmax       & 0.8650±0.0001  & 	0.8894±0.005         \\
With Sigmoid Layer      & 0.8296±0.00024 & 0.8322±0.004      \\
With Softmax Layer      & 0.8188±0.0001 & 0.8093±0.00011      \\
\hline
\end{tabular}
\caption{We observe that performance reduces when we introduce the sigmoid or softmax layer.}
\label{softmax}
\end{table*}

\subsection{Change in architecture}
\label{change-arch}

\subsubsection{Changing Transformer and keeping the CNN same}

From Table \ref{differentTrasnformer} and \ref{differentTrasnformer_results}, we observed that CASS-trained ViT Transformer with the same CNN consistently gained approximately 4.7\% over its
supervised counterpart. Furthermore, from Table \ref{differentTrasnformer_results}, we observed that although ViT L/16 performs better than ViT B/16 on ImageNet ( \cite{rw2019timm}'s results), we observed that the trend is opposite on the autoimmune dataset. Hence, the supervised performance of architecture must be considered before pairing it with CASS.

\begin{table}[!htb]
\centering
\begin{tabular}{lll}
\hline
 Transformer            & CNN F1 Score  & Transformer F1 Score \\
\hline
 ViT Base/16   & 0.8650±0.001 & 0.8894± 0.005       \\
ViT Large/16  & 0.8481±0.001 & 0.853±0.004       \\
\hline
\end{tabular}
\caption{In this table, we show the performance of CASS for ViT large/16 with ResNet-50 and ViT base/16 with ResNet-50. We observed that CASS-trained Transformers, on average, performed 4.7\% better than their supervised counterparts.}
\label{differentTrasnformer}
\end{table}

\begin{table}[!htb]
\centering
\begin{tabular}{ll}
\hline
\multicolumn{1}{c}{Architecture} & Testing F1 Score \\
\hline
ResnNet-50                       & 0.83895±0.007     \\
ViT Base/16                      & 0.8420±0.009     \\
ViT large/16                     & 0.80495±0.0077 \\
\hline
\end{tabular}
\caption{Supervised performance of ViT family on the autoimmune dataset. We observed that as opposed to ImageNet performance, ViT large/16 performs worse than ViT Base/16 on the autoimmune dataset.}
\label{differentTrasnformer_results}
\end{table}

We keep the CNN constant for this experiment and study the effect of changing the Transformer. For this experiment, we use ResNet as our choice of CNN and ViT base and large Transformers with 16 patches. Additionally, we also report performance for DeiT-B \cite{touvron2020training} with ResNet-50. We report these results in Table \ref{transformer_bmri}. Similar to Table \ref{differentTrasnformer}, we observe that by changing Transformer from ViT Base to Large while keeping the number of tokens the same at 16, performance drops. Additionally, for approximately the same size, out of DeiT base and ViT base Transformers, DeiT performs much better than ViT base. 

\begin{table*}[t]
\centering
\begin{tabular}{clll}
\hline
CNN                                                                           & Transformer            & CNN F1 Score  & Transformer F1 Score \\
\hline
\multirow{3}{*}{\begin{tabular}[c]{@{}c@{}}ResNet-50\\ (25.56M)\end{tabular}} & DEiT Base/16 (86.86M)     & 0.9902±0.0025   & 0.9844±0.0048     \\
& ViT Base/16 (86.86M)   & 0.9909±0.0032 & 0.9279± 0.0213       \\
                                                                              & ViT Large/16 (304.72M) & 0.98945±2.45e-5 & 0.8896±0.0009       \\
                                                                              
\hline
\end{tabular}
\caption{For the same number of Transformer parameters, DEiT-base with ResNet-50 performed much better than ResNet-50 with ViT-base. The difference in their CNN arm is ~0.10\%. On ImageNet DEiT-base has a top1\% accuracy of 83.106 while ViT-base has an accuracy of 86.006. We use both Transformers with 16 patches. [ResNet-50 has an accuracy of 80.374] }
\label{transformer_bmri}
\end{table*}

\subsubsection{Changing CNN and keeping the Transformer same}

Table \ref{sameTransformer_atm} and \ref{cnn_atm} we observed that similar to changing Transformer while keeping the CNN same, CASS-trained CNNs gained an average of 3\% over their supervised counterparts. ResNet-200 \cite{rw2019timm} doesn't have ImageNet initialization hence using random initialization. 

\begin{table*}[!htb]
\centering
\begin{tabular}{llll}
\hline
CNN                           & Transformer                  & \multicolumn{2}{l}{100\% Label Fraction} \\
                              &                              & CNN F1 score     & Transformer F1 score   \\
\hline
\multicolumn{1}{c}{ResNet-18 (\textbf{11.69M})} & \multirow{3}{*}{ViT Base/16 (\textbf{86.86M})} & 0.8674±4.8e-5     & 0.8773±5.29e-5           \\
ResNet-50 (\textbf{25.56M})                     &                              & 0.8680±0.001   & 0.8894± 0.0005         \\
ResNet-200 (\textbf{64.69M})                    &                              & 0.8517±0.0009     & 0.874±0.0006     \\
\hline

\end{tabular}
\caption{F1 metric comparison between the two arms of CASS trained over 100 epochs, following the protocols and procedure listed in Appendix \ref{ssl-train}. The numbers in parentheses show the parameters learned by the network. We use \cite{rw2019timm} implementation of CNN and transformers, with ImageNet initialization except for ResNet-200.}
\label{sameTransformer_atm}
\end{table*}

\begin{table}[!htb]
\centering
\begin{tabular}{ll}
\hline
\multicolumn{1}{c}{Architecture} & Testing F1 Score                                        \\
\hline
ResnNet-18                       & 0.8499±0.0004                                           \\
ResnNet-50                       & 0.83895±0.007                                            \\
ResnNet-200                      & 0.833±0.0005  \\
\hline
\end{tabular}
\caption{Supervised performance of the ResNet CNN family on the autoimmune dataset.}
\label{cnn_atm}
\end{table}

For this experiment, we use the ResNet family of CNNs and ViT base/16 as our Transformer. We use ImageNet initialization for ResNet 18 and 50, while random initialization for ResNet-200 (As Timm's library doesn't have an ImageNet initialization). We present these results in Table \ref{differentcnn_bmri}. We observed that an increase in the performance of ResNet correlates to an increase in the performance of the Transformer, implying that there is information transfer between the two.

\begin{table*}[!htb]
\centering
\begin{tabular}{llll}
\hline
CNN                           & Transformer                  & \multicolumn{2}{l}{100\% Label Fraction} \\
                              &                              & CNN F1 score     & Transformer F1 score   \\
\hline
\multicolumn{1}{c}{ResNet-18 (\textbf{11.69M})} & \multirow{3}{*}{ViT Base/16 (\textbf{86.86M})} & 0.9913±0.002     & 0.9801±0.007           \\
ResNet-50 (\textbf{25.56M})                     &                              & 0.9909±0.0032    & 0.9279± 0.0213         \\
ResNet-200 (\textbf{64.69M})                    &                              & 0.9898±0.005     & 0.9276±0.017     \\
\hline

\end{tabular}
\caption{F1 metric comparison between the two arms of CASS trained over 100 epochs, following the protocols and procedure listed in Appendix \ref{ssl-train} and \ref{sup-train}. The numbers in parentheses show the parameters learned by the network. We use \cite{rw2019timm} implementation of CNN and transformers, with ImageNet initialization except for ResNet-200.}
\label{differentcnn_bmri}
\end{table*}

\subsubsection{Using CNN in both arms}

We have experimented using a CNN and a Transformer in CASS on the brain tumor MRI classification dataset. In this section, we present results for using two CNNs in CASS. We pair ResNet-50 with DenseNet-161. We observe that both CNNs fail to reach the benchmark set by ResNet-50 and ViT-B/16 combination. Although training the ResNet-50-DenseNet-161 pair takes 5 hours 24 minutes, less than the 7 hours 11 minutes taken by the ResNet-50-ViT-B/16 combination to be trained with CASS. We compare these results in Table \ref{bothcnn}.

\begin{table*}[!htb]
\centering
\begin{tabular}{clll}
\hline
CNN                        & \begin{tabular}[c]{@{}l@{}}Architecture in\\  $arm_2$\end{tabular} & F1 Score of ResNet-50 arm  & F1 Score of $arm_2$ \\
\hline
\multirow{2}{*}{ResNet-50} & ViT Base/16                                                              & 0.9909±0.0032 & 0.9279± 0.0213            \\
                           & DenseNet-161                                                             & 0.9743±8.8e-5 & 0.98365±9.63e-5    \\
\hline
\end{tabular}
\caption{We observed that for the ResNet-50-DenseNet-161 pair, we train two CNNs instead of one in our standard setup of CASS. Furthermore, none of these CNNs could match the performance of ResNet-50 trained with the ResNet-50-ViT base/16 combination. Hence, by adding a Transformer-CNN combination, we transfer information between the two architectures that would have been missed otherwise.}
\label{bothcnn}
\end{table*}

\subsubsection{Using Transformer in both arms}

Similar to the above section, we use a Transformer-Transformer combination instead of a CNN-Transformer combination. We use Swin-Transformer patch-4/window-12 \cite{Liu_2021_ICCV} alongside ViT-B/16 Transformer. We observe that the performance for ViT/B-16 improves by around ~1.3\% when we use Swin Transformer. However, this comes at a computational cost. The swin-ViT combination took 10 hours to train as opposed to 7 hours and 11 minutes taken by the ResNet-50-ViT-B/16 combination to be trained with CASS. Even with the increased time to train the Swin-ViT combination, it is still almost 50\% less than DINO.
We present these results in Table \ref{bothT}.

\begin{table*}[!htb]
\centering
\begin{tabular}{clll}
\hline
\begin{tabular}[c]{@{}c@{}}Architecture in\\  $arm_1$\end{tabular} & Transformer                   & F1 Score of $arm_1$  & F1 Score of ViT-B/16 arm \\
\hline
\multicolumn{1}{l}{ResNet-50}                                            & \multirow{2}{*}{ViT Base/16} & 0.9909±0.0032  & 0.9279± 0.0213            \\
Swin Transformer                                                          &                              & 0.9883±1.26e-5 & 0.94±8.12e-5     \\
\hline
\end{tabular}
\caption{We present the results for using Transformers in both arms and compare the results with the CNN-Transformer combination.}
\label{bothT}
\end{table*}

\section{Result Analysis}

\subsection{Time complexity analysis}
In Section \ref{time}, we observed that CASS takes 69\% less time than DINO. This reduction in time could be attributed to the following reasons:
\begin{enumerate}
    \item In DINO, augmentations are applied twice as opposed to just once in CASS. Furthermore, per application, CASS uses fewer augmentations than DINO.
    \item Since the architectures used are different, there is no scope for parameter sharing between them. A major chunk of time is saved by updating the two architectures after each epoch instead of re-initializing architectures with lagging parameters.  
\end{enumerate}
\subsection{Qualitative analysis}
\label{Qual}
In this section, we study the feature maps of CNN and attention maps of Transformers trained using CASS and supervised techniques to expand our study. To reinstate, based on the study by \cite{Raghu2021DoVT}, since CNN and Transformer extract different kinds of features from the same input, combing the two would help us create enriched positive pairs for self-supervised learning. In doing so, we would transfer embedded information between the two architectures. We have already seen that this yields better performance in most cases over four different datasets and with three different label fractions. In this section, we study this gain qualitatively with the help of feature maps and class attention maps. 
Based on our study, we observed that CASS-trained Transformers have a more local understanding of the image and hence a more connected attention map than purely-supervised Transformers.

\subsection{Feature maps}
In this section, we study the feature maps from the first five layers of the ResNet-50 model trained with CASS and supervision. We extracted feature maps after the Conv2d layer of ResNet-50. We present the extracted features in Figure \ref{fig:fmaps}. We observed that CASS-trained CNN could retain much more detail about the input image than supervised CNN.

\begin{figure}
    \centering
    \includegraphics[width=0.5\linewidth]{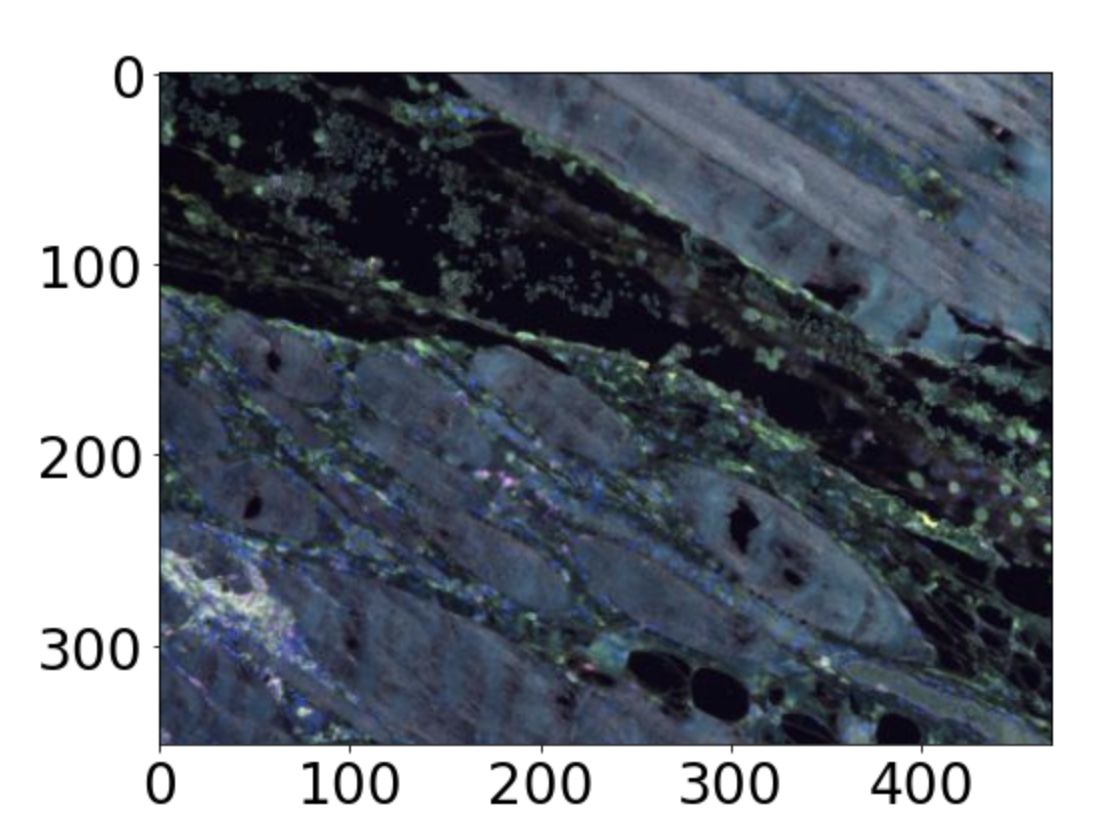}
    \caption{Sample image used from the test set of the autoimmune dataset.}
    \label{fig:sample_image}
\end{figure}

\begin{figure}[!ht]
    \centering
    \includegraphics[width=1\linewidth]{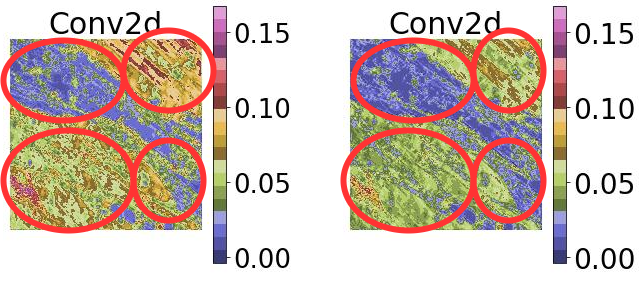}
    \caption{This figure shows the feature map extracted after the first Conv2d layer of ResNet-50 for CASS (on the left) and supervised CNN (on the right). The color bar shows the intensity of pixels retained. From the four circles, it is clear that CASS-trained CNN can retain more intricate details about the input image (Figure \ref{fig:sample_image}) more intensely so that they can be propagated through the architecture and help the model learn better representations as compared to the supervised CNN. We study the same feature map in detail for the first five layers after Conv2d in Figure \ref{fig:fmaps}.}
    
    \label{fig:fmapssingle}
\end{figure}

\begin{figure*}[t]
    \includegraphics[width=0.9\linewidth]{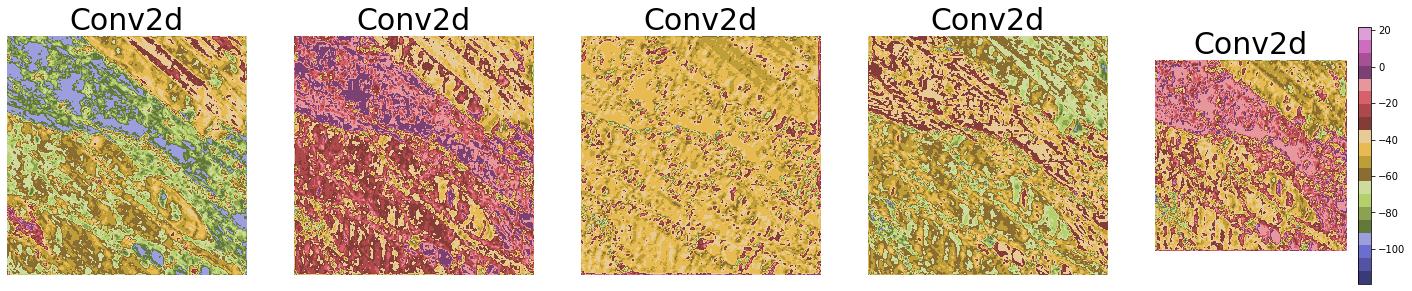}
    \includegraphics[width=0.9\linewidth]{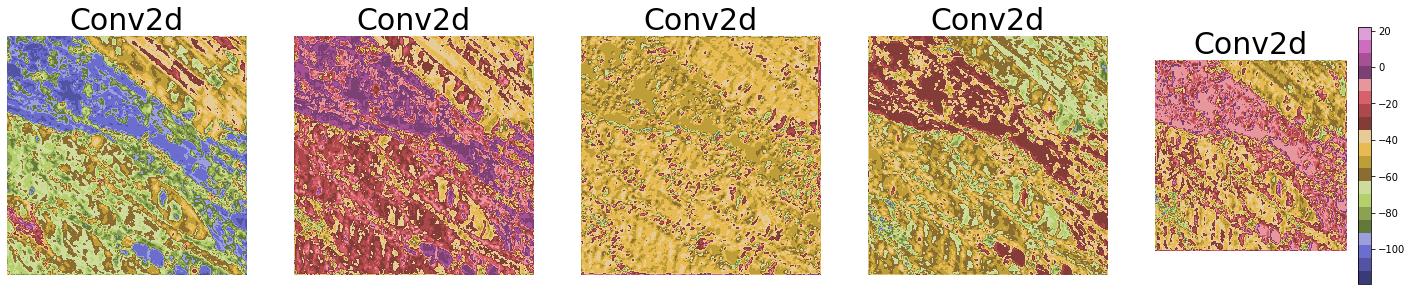}
    \caption{ At the top, we have features extracted from the top 5 layers of supervised ResNet-50, while at the bottom, we have features extracted from the top 5 layers of CASS-trained ResNet-50. We supplied both networks with the same input ( shown in Figure~\ref{fig:sample_image}). }
    \label{fig:fmaps}
\end{figure*}

\subsection{Class attention maps}
\label{attn-map-appendix}

This section will explore the average class attention maps for all four datasets. We studied the attention maps averaged over 30 random samples for autoimmune, dermofit, and brain MRI datasets. Since the ISIC 2019 dataset is highly unbalanced, we averaged the attention maps over 100 samples so that each class may have an example in our sample. We maintained the same distribution as the test set, which has the same class distribution as the overall training set. We observed that CASS-trained Transformers were better able to map global and local connections due to Transformers' ability to map global dependencies and by learning features sensitive to translation equivariance and locality from CNN. This helps the Transformer learn features and local patterns that it would have missed.

\subsubsection{Autoimmune dataset}
\begin{figure}[!ht]
    \centering
    \includegraphics[width=0.4\linewidth]{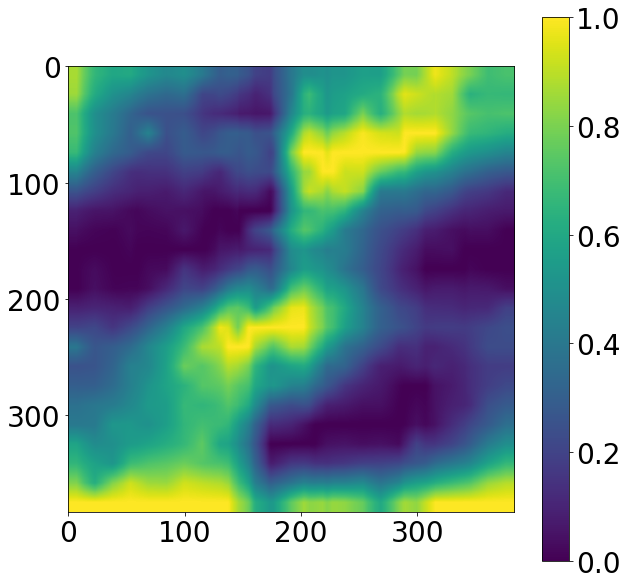}
    \includegraphics[width=0.4\linewidth]{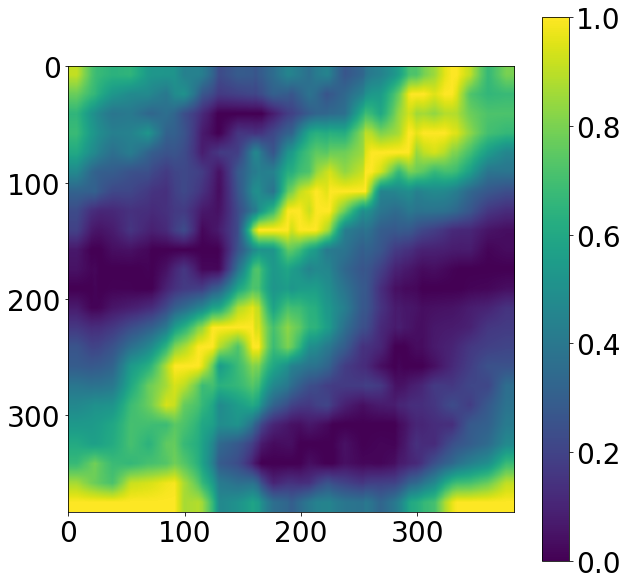}
   
    \caption{To ensure the consistency of our study, we studied average attention maps over 30 sample images from the autoimmune dataset. The left image is the overall attention map averaged over 30 samples for the supervised Transformer, while the one on the right is for CASS pretrained Transformer (both after finetuning with 100\% labels).}
    \label{fig:colton_avg}
\end{figure}

We study the class attention maps averaged over 30 test samples for the autoimmune dataset in Figure \ref{fig:colton_avg}. We observed that the CASS-trained Transformer has much more attention in the center than the supervised Transformer. This extra attention could be attributed to a Transformer on its own inability to map out due to the information transfer from CNN. Another feature to observe is that the attention map of the CASS-trained Transformer is much more connected than that of a supervised Transformer.

    \begin{figure}[!hb]
    \centering
    \includegraphics[width=0.4\linewidth]{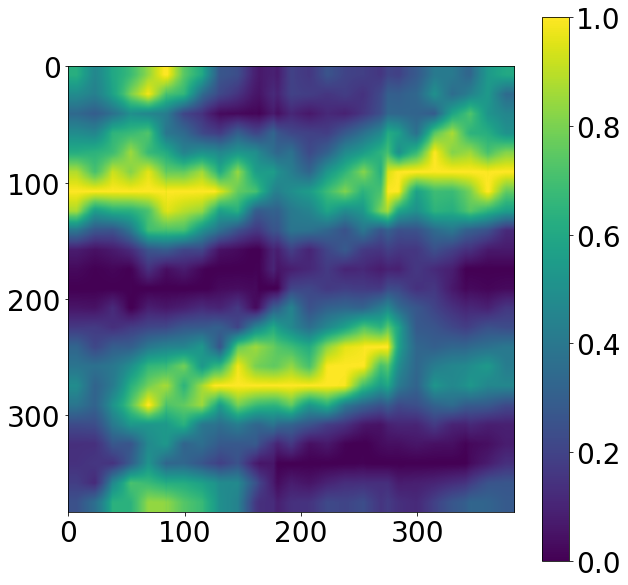}
    \includegraphics[width=0.4\linewidth]{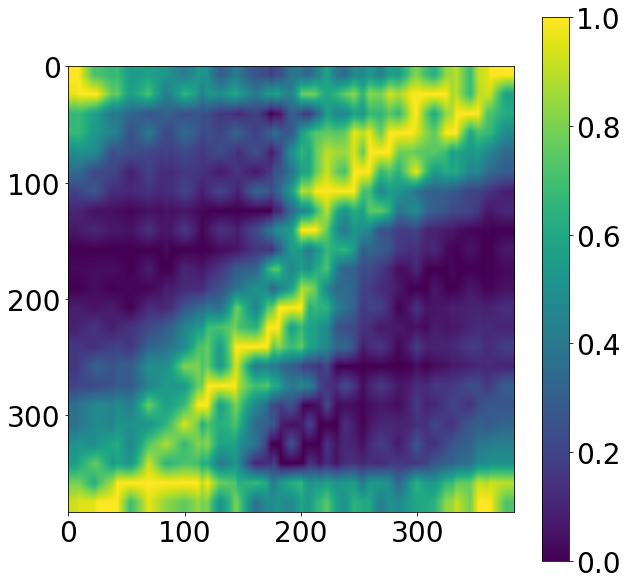}
    
    \caption{Class attention maps averaged over 30 samples of the dermofit dataset for supervised Transformer (on the left), and CASS pretrained Transformer (on the right). Both after finetuning with 100\% labels.}
    
    \label{fig:dermofit_attn}
\end{figure}

\subsubsection{Dermofit dataset} We present the average attention maps for the dermofit dataset in Figure \ref{fig:dermofit_attn}. We observed that the CASS-trained Transformer can pay much more attention to the center part of the image. Furthermore, the attention map of the CASS-trained Transformer is much more connected than the supervised Transformer. So, overall with CASS, the Transformer is not only able to map long-range dependencies which are innate to Transformers but is also able to make more local connections with the help of features sensitive to translation equivariance and locality from CNN.
\begin{figure}[!ht]
    \centering
    \includegraphics[width=0.4\linewidth]{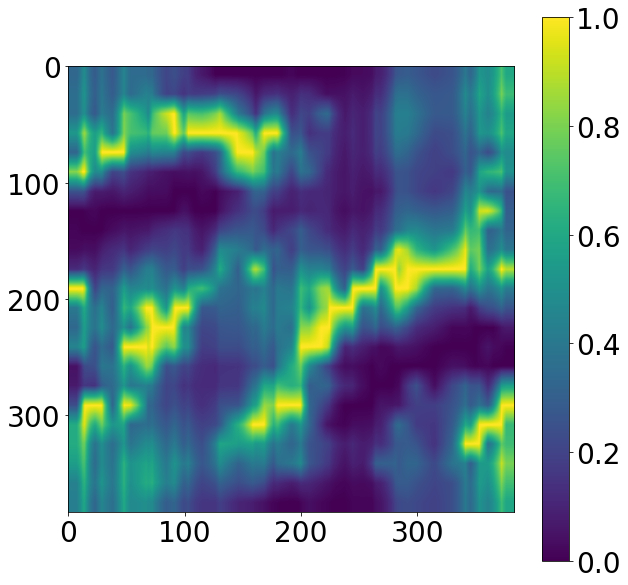}
    \includegraphics[width=0.4\linewidth]{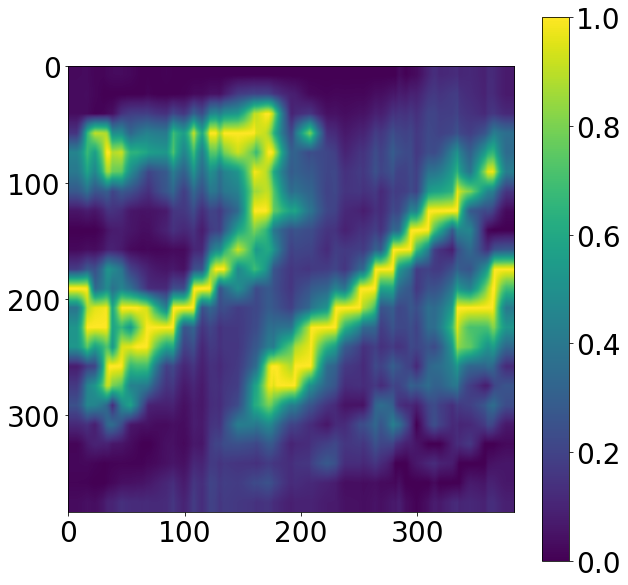}
    \caption{Class attention maps averaged over 30 samples of the brain tumor MRI classification dataset for supervised Transformer (on the left), and CASS pretrained Transformer (on the right). Both after finetuning with 100\% labels.}
    
    \label{fig:bmri_attn}
\end{figure}
    

\subsubsection{Brain tumor MRI classification dataset} We present the average class attention maps results in Figure \ref{fig:bmri_attn}. We observed that a CASS-trained Transformer could better capture long and short-range dependencies than a supervised Transformer. Furthermore, we observed that a CASS-trained Transformer's attention map is much more centered than a supervised Transformer's. From Figure \ref{fig:bmri}, we can observe that most MRI images are center localized, so having a more centered attention map is advantageous in this case.   

\subsubsection{ISIC 2019 dataset} 
\begin{figure}[!ht]
    \centering
    \includegraphics[width=0.4\linewidth]{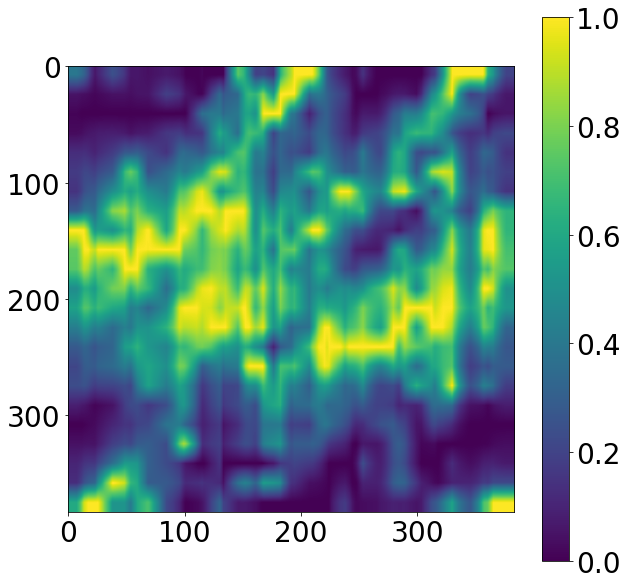}
    \includegraphics[width=0.4\linewidth]{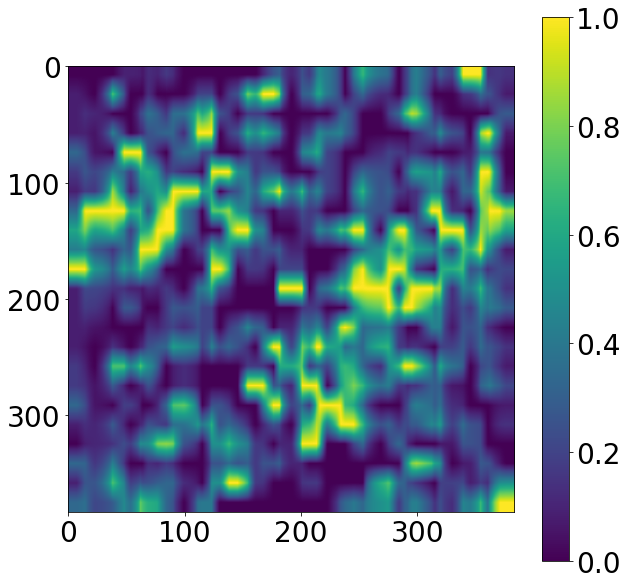}
    \caption{Class attention maps averaged over 100 samples from the ISIC-2019 dataset for the supervised Transformer (on the left) and CASS-trained Transformer (on the right). Both after finetuning with 100\% labels.}
    
    \label{fig:isic_attn}
\end{figure}

The ISIC-2019 dataset is one of the most challenging datasets out of the four datasets. ISIC 2019 consists of images from the HAM10000 and BCN\_20000 datasets \cite{cassidy2022analysis, Gessert2020SkinLC}. For the HAM1000 dataset, it isn't easy to classify between 4 classes (melanoma and melanocytic nevus), (actinic keratosis, and benign keratosis). HAM10000 dataset contains images of size 600×450, centered and cropped around the lesion. Histogram corrections have been applied to only a few images. The BCN\_20000 dataset contains images of size 1024×1024. This dataset is particularly challenging as many images are uncropped, and lesions are in difficult and uncommon locations. Hence, in this case, having more spread-out attention maps would be advantageous instead of a more centered one. From Figure \ref{fig:isic_attn}, we observed that a CASS-trained Transformer has a lot more spread attention map than a supervised Transformer. Furthermore, a CASS-trained Transformer can also attend the corners far better than a supervised Transformer.  

From Figures \ref{fig:colton_avg}, \ref{fig:dermofit_attn}, \ref{fig:bmri_attn} and \ref{fig:isic_attn}, we observed that in most cases, the supervised Transformer had spread out attention, while the CASS trained Transformer has a more "connected."
attention map. This is primarily because of local-level information transfer from CNN. Hence we could add some more image-level intuition, with the help of CNN, to the Transformer that it would have rather missed on its own.

\subsubsection{Choice of Datasets}
\label{ds-choice}
\begin{figure}[!h]
    \centering
    \includegraphics[width=0.4\linewidth]{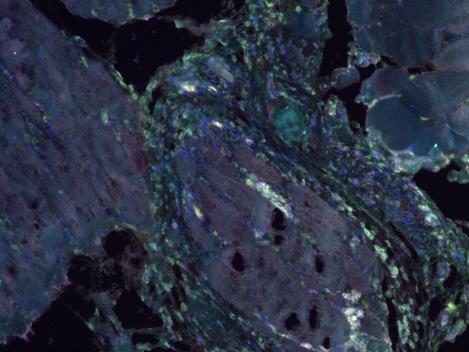}
    \includegraphics[width=0.4\linewidth]{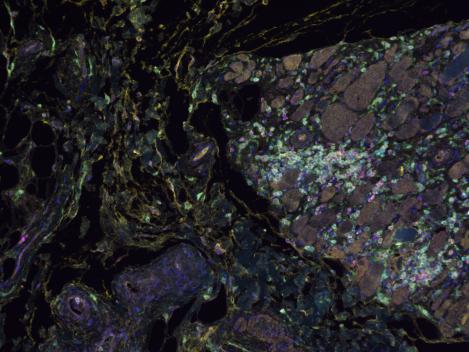}
    \caption{Sample of autofluorescence slide images from the muscle biopsy of patients with dermatomyositis - a type of autoimmune disease.}
    \label{fig:autoimmune}
\end{figure}

\begin{figure}[!htb]
    \centering  
    \includegraphics[width=0.3\linewidth]{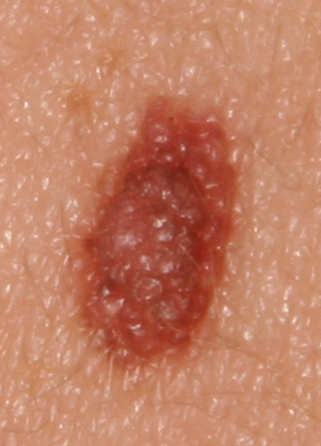}
    \includegraphics[width=0.3\linewidth]{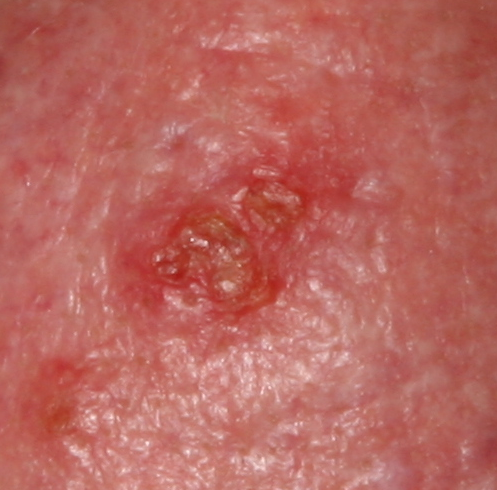}
    \includegraphics[width=0.3\linewidth]{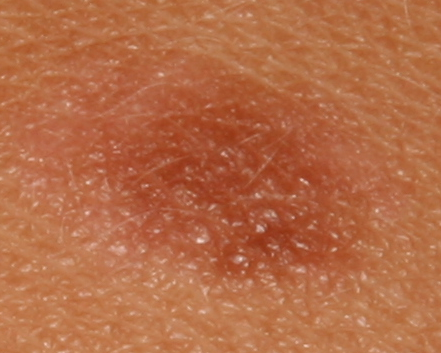}
     \hspace{1cm}
    \caption{Sample images from the Dermofit dataset.}
    \label{fig:dermofit}
    \end{figure}

\begin{figure}[!htb]
     \centering
    \includegraphics[width=0.25\linewidth]{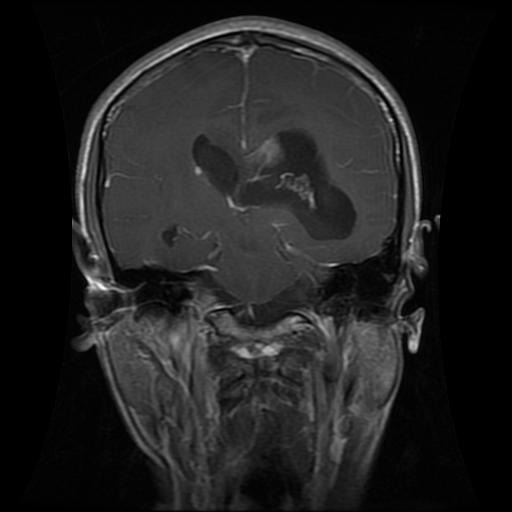}
    \includegraphics[width=0.25\linewidth]{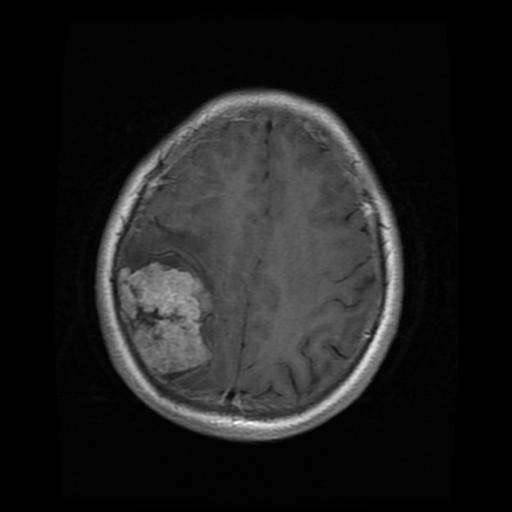}
    \includegraphics[width=0.25\linewidth]{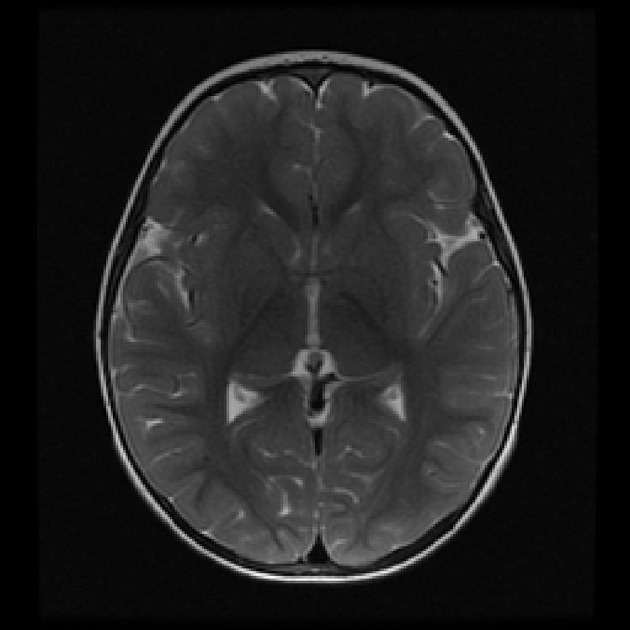}
     \hspace{1cm}
    \caption{Sample images of brain tumor MRI dataset, Each image corresponds to a prediction class in the data set glioma (Left), meningioma (Center), and No tumor (Right)  }
    \label{fig:bmri}
    \end{figure}

\begin{figure}[!htb]
    \centering  
    \includegraphics[width=0.25\linewidth]{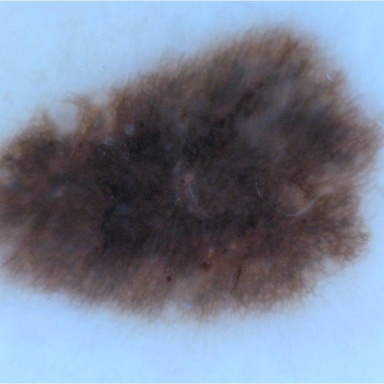}
    \includegraphics[width=0.25\linewidth]{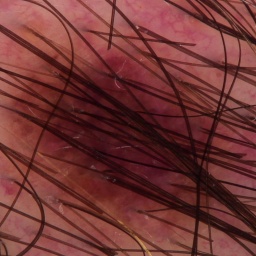}
    \includegraphics[width=0.25\linewidth]{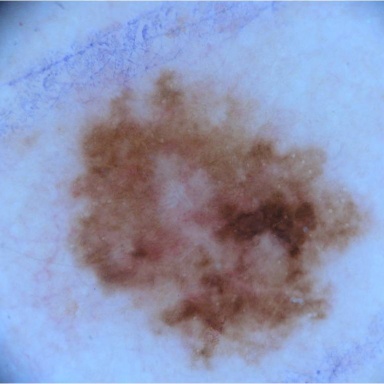}
     \hspace{1cm}
    \caption{Sample images from the ISIC-2019 challenge dataset.}
    \label{fig:isic}
    \end{figure}

We chose four medical imaging datasets with diverse sample sizes ranging from 198 to 25,336, along with diverse modalities to study the performance of existing self-supervised techniques and CASS.
Most of the existing self-supervised techniques have been studied on million image datasets like ImageNet, but medical imaging datasets, on average, are much smaller than a million images. Furthermore, they are usually imbalanced, and some existing self-supervised techniques rely on batch statistics, making them learn skewed representations. We also include a dataset of emerging and underrepresented diseases with only a few hundred samples, the autoimmune dataset in our case (198 images). To the best of our knowledge, no existing literature studies the effect of self-supervised learning on such a small dataset. Furthermore, we chose the dermofit dataset because all the images are taken using an SLR camera, and no two images are the same size. Image size in dermofit varies from 205×205 to 1020×1020. MRI images constitute a large part of medical imaging; hence we included this dataset in our study. So, to incorporate them into our study, we had the Brain tumor MRI classification dataset. Furthermore, it is our study's only black-and-white dataset; the other three datasets are RGB.
The ISIC 2019 is a unique dataset containing multiple pairs of hard-to-classify classes (Melanoma - melanocytic nevus and actinic keratosis - benign keratosis) and different image sizes - out of which only a few have been prepossessed. It is a highly imbalanced dataset containing samples with lesions in difficult and uncommon locations. To give an idea about the images used in our experiments, we provide sample images from the four datasets used in our experimentation in Figures \ref{fig:autoimmune}, \ref{fig:dermofit}, \ref{fig:bmri} and \ref{fig:isic}. 

\subsection{Self-supervised pretraining}
\label{ssl-train}
\subsubsection{Protocols}
\begin{itemize}
    \item Self-supervised learning was only done on the training data, not the validation data. We used \url{https://github.com/PyTorchLightning/pytorch-lightning} to set the pseudo-random number generators in PyTorch, NumPy, and (python.random).

   \item We ran training over five seed values and reported mean results with variance in each table. We didn't perform a seed value sweep to extract any more performance \cite{picard2021torch}.
   
   \item For MAE, BYOL, and DINO implementation, we use Phil Wang's implementation: \url{https://github.com/lucidrains/vit-pytorch} and \url{https://github.com/lucidrains/vit-pytorch}.
   
   \item For the implementation of CNNs and Transformers, we use Timm's library \cite{rw2019timm}.
   
   \item The goal of self-supervision is to provide better initialization; for our set of experiments, we used ImageNet initialization \cite{deng2009imagenet}; because ImageNet initializations provide faster convergence and better weight scaling \cite{raghu2019transfusion}. In the case of CASS, the differences in ImageNet learned representations also act as a scaffolding to avoid collapse.   

   \item After pertaining, an end-to-end finetuning of the pre-trained model was done using x\% labeled data. Where x was either 1, 10, or 100. When fine-tuned with x\% labeled data, the pre-trained model was fine-tuned only on x\% data points with corresponding labels. 
\end{itemize}

\subsubsection{Augmentations}

\begin{itemize}

\item Resizing: Resize input images to 384×384 with bilinear interpolation.
\item Color jittering: change the brightness, contrast, saturation, and hue of an image or apply random perspective with a given probability. We set the degree of distortion to 0.2 (between 0 and 1) and use bilinear interpolation with an application probability of 0.3.
\item Color jittering or applying the random affine transformation of the image, keeping center invariant with degree 10, with an application probability of 0.3.
\item Horizontal and Vertical flip. Each with an application probability of 0.3.
\item Channel normalization with a mean (0.485, 0.456, 0.406) and standard deviation (0.229, 0.224, 0.225).

\end{itemize}

\subsubsection{Hyper-parameters}

\begin{itemize}
    \item Optimization: We use stochastic weighted averaging over Adam optimizer with learning rate (LR) set to 1e-3 for both CNN and vision transformer (ViT). This is a shift from SGD, which is usually used for CNNs.
    
    \item Learning Rate: Cosine annealing learning rate is used with 16 iterations and a minimum learning rate of 1e-6. Unless mentioned otherwise, this setup was trained over 100 epochs. These were then used as initialization for the downstream supervised learning. The standard batch size is 16.
    
\end{itemize}

\subsection{Supervised training}
\label{sup-train}
\subsubsection{Augmentations}
We use the same set of augmentations used in self-supervised pretraining.
\subsubsection{Hyper-parameters}
\begin{itemize}
    \item We use Adam optimizer with lr set to 3e-4 and a cosine annealing learning schedule.
    \item Since all medical datasets have a class imbalance, we address it by using focal loss \cite{Lin2017FocalLF} as our choice of the loss function with the alpha value set to 1 and the gamma value to 2. In our case, it uses minimum-maximum normalized class distribution as class weights for focal loss.
    \item We train for 50 epochs. We also use a five epoch patience on validation loss to check for early stopping. This downstream supervised learning setup is kept the same for CNN and Transformers.
\end{itemize}

We repeat all the experiments with different seed values five times and then present the average results in all the tables.

\subsection{Description of Metrics}
\label{metrics}
After performing downstream fine-tuning on the four datasets under consideration, we analyze the CASS, DINO, and Supervised approaches on specific metrics for each dataset. The choice of this metric is either from previous work or as defined by the dataset provider. For the Autoimmune dataset, Dermofit, and Brain MRI classification datasets based on prior works, we use the F1 score as our metric for comparing performance, which is defined as $F1 = \frac{2*Precision*Recall}{Precision+Recall} = \frac{2*TP}{2*TP+FP+FN}$

For the ISIC-2019 dataset, as mentioned by the competition organizers, we used the recall score as our comparison metric, which is defined as $Recall = \frac{TP}{TP+FN}$

For the above two equations, TP: True Positive, TN: True Negative, FP: False Positive, and FN: False Negative.





\end{document}